\documentclass[10pt,twocolumn,letterpaper]{article}

\usepackage[pagenumbers]{cvpr} %

\usepackage{graphicx}
\usepackage{amsmath}
\usepackage{amssymb}
\usepackage{booktabs}
\usepackage[T1]{fontenc}
\usepackage{caption}
\usepackage{subcaption}
\makeatletter
\@namedef{ver@everyshi.sty}{}
\makeatother
\usepackage{tikz}
\usetikzlibrary{spy,calc}

\usepackage[pagebackref,breaklinks,colorlinks]{hyperref}

\usepackage[capitalize]{cleveref}
\crefname{section}{Sec.}{Secs.}
\Crefname{section}{Section}{Sections}
\Crefname{table}{Table}{Tables}
\crefname{table}{Tab.}{Tabs.}

\def\cvprPaperID{5654} %
\def\confName{CVPR}
\def\confYear{2023}

\newif\ifblackandwhitecycle
\gdef\patternnumber{0}

\pgfkeys{/tikz/.cd,
    zoombox paths/.style={
        draw=orange,
        very thick
    },
    black and white/.is choice,
    black and white/.default=static,
    black and white/static/.style={ 
        draw=white,   
        zoombox paths/.append style={
            draw=white,
            postaction={
                draw=black,
                loosely dashed
            }
        }
    },
    black and white/static/.code={
        \gdef\patternnumber{1}
    },
    black and white/cycle/.code={
        \blackandwhitecycletrue
        \gdef\patternnumber{1}
    },
    black and white pattern/.is choice,
    black and white pattern/0/.style={},
    black and white pattern/1/.style={    
            draw=white,
            postaction={
                draw=black,
                dash pattern=on 2pt off 2pt
            }
    },
    black and white pattern/2/.style={    
            draw=white,
            postaction={
                draw=black,
                dash pattern=on 4pt off 4pt
            }
    },
    black and white pattern/3/.style={    
            draw=white,
            postaction={
                draw=black,
                dash pattern=on 4pt off 4pt on 1pt off 4pt
            }
    },
    black and white pattern/4/.style={    
            draw=white,
            postaction={
                draw=black,
                dash pattern=on 4pt off 2pt on 2 pt off 2pt on 2 pt off 2pt
            }
    },
    zoomboxarray inner gap/.initial=5pt,
    zoomboxarray columns/.initial=2,
    zoomboxarray rows/.initial=2,
    subfigurename/.initial={},
    figurename/.initial={zoombox},
    zoomboxarray/.style={
        execute at begin picture={
            \begin{scope}[
                spy using outlines={%
                    zoombox paths,
                    width=\imagewidth / \pgfkeysvalueof{/tikz/zoomboxarray columns} - (\pgfkeysvalueof{/tikz/zoomboxarray columns} - 1) / \pgfkeysvalueof{/tikz/zoomboxarray columns} * \pgfkeysvalueof{/tikz/zoomboxarray inner gap} -\pgflinewidth,
                    height=\imageheight / 2 / \pgfkeysvalueof{/tikz/zoomboxarray rows} - (\pgfkeysvalueof{/tikz/zoomboxarray rows} - 1) / \pgfkeysvalueof{/tikz/zoomboxarray rows} * \pgfkeysvalueof{/tikz/zoomboxarray inner gap}-\pgflinewidth,
                    magnification=3,
                    every spy on node/.style={
                        zoombox paths
                    },
                    every spy in node/.style={
                        zoombox paths
                    }
                }
            ]   
        },
        execute at end picture={
            \end{scope}
            \node at (image.north) [anchor=north,inner sep=0pt] {\subcaptionbox{\label{\pgfkeysvalueof{/tikz/figurename}-image}}{\phantomimage}};
            \node at (zoomboxes container.north) [anchor=north,inner sep=0pt] {\subcaptionbox{\label{\pgfkeysvalueof{/tikz/figurename}-zoom}}{\phantomimage}};
     \gdef\patternnumber{0}
        },
        spymargin/.initial=0.5em,
        zoomboxes xshift/.initial=1,
        zoomboxes right/.code=\pgfkeys{/tikz/zoomboxes xshift=1},
        zoomboxes left/.code=\pgfkeys{/tikz/zoomboxes xshift=-1},
        zoomboxes yshift/.initial=0,
        zoomboxes above/.code={
            \pgfkeys{/tikz/zoomboxes yshift=1},
            \pgfkeys{/tikz/zoomboxes xshift=0}
        },
        zoomboxes below/.code={
            \pgfkeys{/tikz/zoomboxes yshift=-1},
            \pgfkeys{/tikz/zoomboxes xshift=0}
        },
        caption margin/.initial=4ex,
    },
    adjust caption spacing/.code={},
    image container/.style={
        inner sep=0pt,
        at=(image.north),
        anchor=north,
        adjust caption spacing
    },
    zoomboxes container/.style={
        inner sep=0pt,
        at=(image.north),
        anchor=north,
        name=zoomboxes container,
        xshift=\pgfkeysvalueof{/tikz/zoomboxes xshift}*(\imagewidth+\pgfkeysvalueof{/tikz/spymargin}),
        yshift=\pgfkeysvalueof{/tikz/zoomboxes yshift}*(\imageheight+\pgfkeysvalueof{/tikz/spymargin}+\pgfkeysvalueof{/tikz/caption margin}),
        adjust caption spacing
    },
    calculate dimensions/.code={
        \pgfpointdiff{\pgfpointanchor{image}{south west} }{\pgfpointanchor{image}{north east} }
        \pgfgetlastxy{\imagewidth}{\imageheight}
        \global\let\imagewidth=\imagewidth
        \global\let\imageheight=\imageheight
        \gdef\columncount{1}
        \gdef\rowcount{1}
        
    },
    image node/.style={
        inner sep=0pt,
        name=image,
        anchor=south west,
        append after command={
            [calculate dimensions]
            node [image container,subfigurename=\pgfkeysvalueof{/tikz/figurename}-image] {\phantomimage}
            node [zoomboxes container,subfigurename=\pgfkeysvalueof{/tikz/figurename}-zoom] {\phantomimage}
        }
    },
    color code/.style={
        zoombox paths/.append style={draw=#1}
    },
    connect zoomboxes/.style={
    spy connection path={\draw[draw=none,zoombox paths] (tikzspyonnode) -- (tikzspyinnode);}
    },
    help grid code/.code={
        \begin{scope}[
                x={(image.south east)},
                y={(image.north west)},
                font=\footnotesize,
                help lines,
                overlay
            ]
            \foreach \x in {0,1,...,9} { 
                \draw(\x/10,0) -- (\x/10,1);
                \node [anchor=north] at (\x/10,0) {0.\x};
            }
            \foreach \y in {0,1,...,9} {
                \draw(0,\y/10) -- (1,\y/10);                        \node [anchor=east] at (0,\y/10) {0.\y};
            }
        \end{scope}    
    },
    help grid/.style={
        append after command={
            [help grid code]
        }
    },
}

\newcommand\phantomimage{%
    \phantom{%
        \rule{\imagewidth}{\imageheight}%
    }%
}
\newcommand\zoombox[2][]{
    \begin{scope}[zoombox paths]
        \pgfmathsetmacro\xpos{
            (\columncount-1)*(\imagewidth / \pgfkeysvalueof{/tikz/zoomboxarray columns} + \pgfkeysvalueof{/tikz/zoomboxarray inner gap} / \pgfkeysvalueof{/tikz/zoomboxarray columns} ) + \pgflinewidth
        }
        \pgfmathsetmacro\ypos{
            (\rowcount-1)*( \imageheight / \pgfkeysvalueof{/tikz/zoomboxarray rows} + \pgfkeysvalueof{/tikz/zoomboxarray inner gap} / \pgfkeysvalueof{/tikz/zoomboxarray rows} ) + 0.5*\pgflinewidth
        }
        \edef\dospy{\noexpand\spy [
            #1,
            zoombox paths/.append style={
                black and white pattern=\patternnumber
            },
            every spy on node/.append style={#1},
            x=\imagewidth,
            y=\imageheight
        ] on (#2) in node [anchor=north west] at ($(zoomboxes container.north west)+(\xpos pt,-\ypos pt)$);}
        \dospy
        \pgfmathtruncatemacro\pgfmathresult{ifthenelse(\columncount==\pgfkeysvalueof{/tikz/zoomboxarray columns},\rowcount+1,\rowcount)}
        \global\let\rowcount=\pgfmathresult
        \pgfmathtruncatemacro\pgfmathresult{ifthenelse(\columncount==\pgfkeysvalueof{/tikz/zoomboxarray columns},1,\columncount+1)}
        \global\let\columncount=\pgfmathresult
        \ifblackandwhitecycle
            \pgfmathtruncatemacro{\newpatternnumber}{\patternnumber+1}
            \global\edef\patternnumber{\newpatternnumber}
        \fi
    \end{scope}
}

\begin{document}

\title{Neural Font Rendering}

\author
{
    Daniel Anderson
    \hspace{6mm} Ariel Shamir
    \hspace{6mm} Ohad Fried
    \\
    Reichman University 
}
 
\maketitle

\newcommand{\betweencellpdf}{\ensuremath{h^c}}
\newcommand{\betweencellpdffine}{\ensuremath{h^f}}
\newcommand{\approxbetweencellpdffine}{\ensuremath{\hat{h}^f}}
\newcommand{\incellpdf}{\ensuremath{f}}
\newcommand{\incellpdfnormalized}{\ensuremath{f'}}

\newcommand{\incellcdf}{\ensuremath{F}}

\newcommand{\totalpdf}{\ensuremath{f_{dd}}}
\newcommand{\totalcdf}{\ensuremath{F_{dd}}}

\newcommand{\ignorethis}[1]{}
\newcommand{\redund}[1]{#1}

\newcommand{\apriori    }     {\textit{a~priori}}
\newcommand{\aposteriori}     {\textit{a~posteriori}}
\newcommand{\perse      }     {\textit{per~se}}
\newcommand{\naive      }     {{na\"{\i}ve}}
\newcommand{\Naive      }     {{Na\"{\i}ve}}
\newcommand{\Identity   }     {\mat{I}}
\newcommand{\Zero       }     {\mathbf{0}}
\newcommand{\Reals      }     {{\textrm{I\kern-0.18em R}}}
\newcommand{\isdefined  }     {\mbox{\hspace{0.5ex}:=\hspace{0.5ex}}}
\newcommand{\texthalf   }     {\ensuremath{\textstyle\frac{1}{2}}}
\newcommand{\half       }     {\ensuremath{\frac{1}{2}}}
\newcommand{\third      }     {\ensuremath{\frac{1}{3}}}
\newcommand{\fourth     }     {\ensuremath{\frac{1}{4}}}

\newcommand{\Lone} {\ensuremath{L_1}}
\newcommand{\Ltwo} {\ensuremath{L_2}}

\newcommand{\degree} {\ensuremath{^{\circ}}}

\newcommand{\mat        } [1] {{\text{\boldmath $\mathbit{#1}$}}}
\newcommand{\Approx     } [1] {\widetilde{#1}}
\newcommand{\change     } [1] {\mbox{{\footnotesize $\Delta$} \kern-3pt}#1}

\newcommand{\Order      } [1] {O(#1)}
\newcommand{\set        } [1] {{\lbrace #1 \rbrace}}
\newcommand{\floor      } [1] {{\lfloor #1 \rfloor}}
\newcommand{\ceil       } [1] {{\lceil  #1 \rceil }}
\newcommand{\inverse    } [1] {{#1}^{-1}}
\newcommand{\transpose  } [1] {{#1}^\mathrm{T}}
\newcommand{\invtransp  } [1] {{#1}^{-\mathrm{T}}}
\newcommand{\relu       } [1] {{\lbrack #1 \rbrack_+}}

\newcommand{\abs        } [1] {{| #1 |}}
\newcommand{\Abs        } [1] {{\left| #1 \right|}}
\newcommand{\norm       } [1] {{\| #1 \|}}
\newcommand{\Norm       } [1] {{\left\| #1 \right\|}}
\newcommand{\pnorm      } [2] {\norm{#1}_{#2}}
\newcommand{\Pnorm      } [2] {\Norm{#1}_{#2}}
\newcommand{\inner      } [2] {{\langle {#1} \, | \, {#2} \rangle}}
\newcommand{\Inner      } [2] {{\left\langle \begin{array}{@{}c|c@{}}
                               \displaystyle {#1} & \displaystyle {#2}
                               \end{array} \right\rangle}}

\newcommand{\twopartdef}[4]
{
  \left\{
  \begin{array}{ll}
    #1 & \mbox{if } #2 \\
    #3 & \mbox{if } #4
  \end{array}
  \right.
}

\newcommand{\fourpartdef}[8]
{
  \left\{
  \begin{array}{ll}
    #1 & \mbox{if } #2 \\
    #3 & \mbox{if } #4 \\
    #5 & \mbox{if } #6 \\
    #7 & \mbox{if } #8
  \end{array}
  \right.
}

\newcommand{\len}[1]{\text{len}(#1)}

\newlength{\w}
\newlength{\h}
\newlength{\x}

\definecolor{darkred}{rgb}{0.7,0.1,0.1}
\definecolor{darkgreen}{rgb}{0.1,0.6,0.1}
\definecolor{cyan}{rgb}{0.7,0.0,0.7}
\definecolor{otherblue}{rgb}{0.1,0.4,0.8}
\definecolor{maroon}{rgb}{0.76,.13,.28}
\definecolor{burntorange}{rgb}{0.81,.33,0}

\ifdefined\ShowNotes
  \newcommand{\colornote}[3]{{\color{#1}\textbf{#2} #3\normalfont}}
\else
  \newcommand{\colornote}[3]{}
\fi

\newcommand {\todo}[1]{\textcolor{cyan}{TODO: {#1}}}
\newcommand {\ohad}[1]{\textcolor{burntorange}{OF: {#1}}}
\newcommand {\as}[1]{\textcolor{darkgreen}{AS: {#1}}}
\newcommand {\daniel}[1]{\textcolor{otherblue}{DA: {#1}}}

\newcommand {\reqs}[1]{\colornote{red}{\tiny #1}}

\newcommand {\new}[1]{\colornote{red}{#1}}

\newcommand*\rot[1]{\rotatebox{90}{#1}}

\newcommand {\newstuff}[1]{#1}

\newcommand\todosilent[1]{}

\newcommand{\woBGmask}{{w/o~bg~\&~mask}}
\newcommand{\woMask}{{w/o~mask}}

\providecommand{\keywords}[1]
{
  \textbf{\textit{Keywords---}} #1
}

\begin{abstract}
    Recent advances in deep learning techniques and applications have revolutionized artistic creation and manipulation in many domains (text, images, music); however, fonts have not yet been integrated with deep learning architectures in a manner that supports their multi-scale nature. In this work we aim to bridge this gap, proposing a network architecture capable of rasterizing glyphs in multiple sizes, potentially paving the way for easy and accessible creation and manipulation of fonts.
\end{abstract}

\section{Introduction}

\begin{figure}\centering
\begin{tikzpicture}
    [spy using outlines={circle,red,magnification=5,size=1.5cm, connect spies}]
    \node {\pgfimage[height=5.5cm]{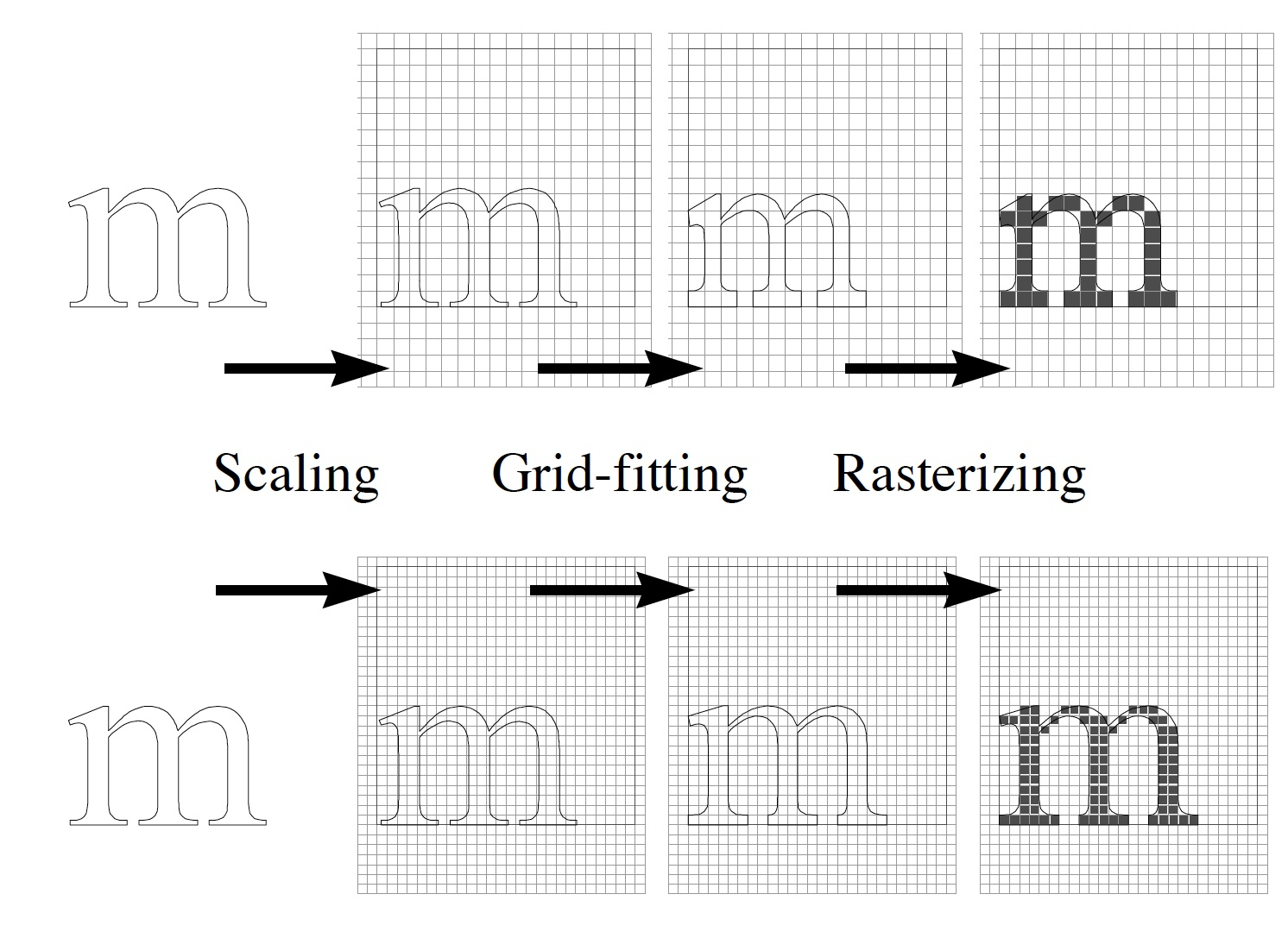}};
    \spy on (-.45,1.02) in node [left] at (-.8,2.35);
    \spy on (1.29,1.02) in node [left] at (.8,2.35);
\end{tikzpicture}
\caption{Traditional conversion of a glyph outline to bitmap involves: (1) scaling: imposing a grid of pixels in a suitable size, (2) grid-fitting using hints: changing the shape of the outline to better fit the grid and preserve typographic constraints, and (3) rasterizing: drawing pixels to create the bitmap (in this case simple black/white). Top: point size 14, bottom: point size 26. Our work investigates the possibility to replace this procedure with an implicit neural representation for glyphs in \emph{any} size.}
\label{fig:rasterization}
\end{figure}

Typography is all around us: 
from large billboards to small text on your mobile device. Digital typefaces (i.e. fonts) have an inherent ability to scale to any desired size and resolution. Each font is composed of a set of glyphs representing characters, numbers, and other symbols.  
The glyph outlines are defined using vector-based curves, and can be drawn in any size via a complex procedure called \emph{glyph rasterization}.
Remarkably, this practice of representing typographic glyphs as explicit boundary curves 
has stayed largely the same since the 1980s~\cite{AdobeType1}. 
Recently, other explicit geometric entities have been augmented by complementary neural implicit representations. 
For instance, NeRF~\cite{Mildenhall20} and its derivatives~\cite{barron2021mipnerf,mildenhall2021rawnerf,barron2022mipnerf360,poole2022dreamfusion} have developed as an alternative representation to the common 3D boundary mesh representation.
Inspired by these recent advances, in this work we study the possibility of representing typographic glyphs as implicit functions, as a possible first step towards a new and complementary representation for fonts.

A major challenge that sets glyph rasterization apart from natural image synthesis and rendering is the requirement to support continuous output resolutions. A font glyph must be rasterized in a legible manner at any bitmap size (above some minimum). Hence, implicit functions as a basic representation seem like a natural choice, as they can be sampled at any desired resolution.
However, the traditional rasterization process of glyphs is far from trivial sampling. Converting the boundary outlines of a glyph to bitmaps entails the following steps (\Cref{fig:rasterization}).
First, scaling to the desired size is performed by overlaying a suitable-size grid of pixels on the outline of the glyph, so that appropriate pixels are chosen to create the bitmap of the glyph. However, simply turning on the pixels that are inside the outline (even with anti-aliasing) creates noticeable artifacts that hurt the output bitmap quality and legibility (e.g., \Cref{hinting-importance-med}). 
To alleviate this, before converting to pixels, the outline of the glyph is modified to better align with the grid of pixels, and only then the pixels are sampled. This process is called grid-fitting and it is governed by what is often called \emph{hints} --- commands that modify the glyph's outline shape depending on the resolution.
Hints preserve typographic constraints such as maintaining widths and distances of glyph elements. Thus, each glyph within a digital font is not defined by a vector-representation outline only, but includes complex hints that support fitting it to all sizes, written in a specialized programming language such as TrueType~\cite{TrueTypeHist} or PostScript~\cite{AdobeType1}. 

\begin{figure}[t!]
    \centering
    \hfill
         \includegraphics{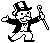} \hfill
         \includegraphics{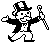} \hfill
         \includegraphics{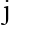} \hfill
         \includegraphics{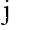} \hfill \hfill
    \caption{Comparison of a single pixel change between a small black \& white image (left) and a glyph bitmap (right). Human perception is much more sensitive to glyphs, due to size, structure and familiarity. Please zoom-in to see differences better.}
    \label{fig:one_px_error}
\end{figure}

In a similar manner, an implicit neural representation of a glyph must support a dynamic change based on the output resolution so that the creation of the output bitmap using point sampling would retain legibility. In contrast to color or  gray-scale images, where difference in the color of one or two pixels may not affect human perception much~\cite{Adverserial-15}, the human eye is extremely sensitive to small errors in gray-scale glyph bitmaps (\Cref{fig:one_px_error} compares the prominence of a single pixel error between a small black \& white image and a glyph image). For this reason, the creation of a neural glyph representation that will produce legible bitmaps in any size is extremely challenging. 

We present a neural representation for glyphs that achieves high quality results compared to a baseline representation. In addition, we present results on encoding multiple typeface glyphs such as multiple weights. These tasks are challenging as they require zero tolerance to even the slightest error. 
There are many recent works applying deep learning and other modern techniques to fonts and glyphs, for example for stylization or font interpolation \cite{Suveeranont09,Campbell14,Upchurch16,Bernhardsson2016,Park2018TypefaceCW,Jiang2019SCFontSC,Xi20,Wang20,BerioStrokestyles2022}. 
However, these works tend to treat the glyphs as images and focus on a single (usually very large) size, eschewing some of the difficulties inherent in actual font manipulation and rasterization. 
Although there are still some artifacts present in the bitmap results, our method takes a first step towards defining a true neural font representation. 
To our knowledge, it is the first to propose a deep learning approach that answers the challenge of a multi-scale font representation. Our code is available at \href{https://github.com/AndersonDaniel/neural-font-rendering}{https://github.com/AndersonDaniel/neural-font-rendering}.

\section{Background on Font Rendering}
\label{sec:font-rendering}

\begin{figure}[t]
  \centering
  \includegraphics[width=\linewidth]{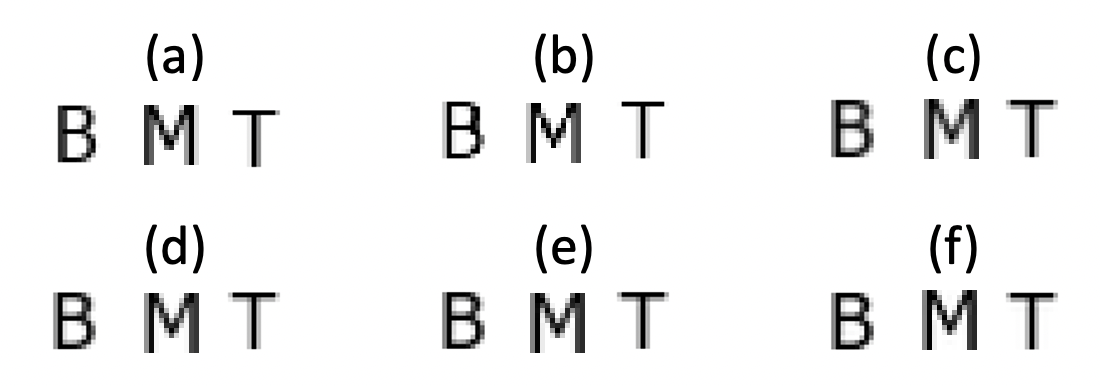}
  \caption{Comparison of several image resizing techniques applied to rendered glyphs.
  (a) Tahoma glyphs B/M/T rendered in pointsize 16.
  (b) - (f) Tahoma glyphs rendered in pointsize 24 and resized to same size of (a) with the following techniques, respectively: (b) nearest neighbor, (c) bilinear interpolation, (d) Hamming interpolation, (e) bicubic interpolation, (f) Lanczos resampling. Differences can be seen between each resizing method vs. rendering directly in the intended size, e.g. in edge sharpness or glyph shape. }
  \label{resize-issues-med}
\end{figure}

Fonts today can include between hundreds to thousands of glyphs, supporting for instance, multiple alphabets, accented versions of letters, punctuation marks, and other symbols. Each glyph can be rendered in a continuous range of sizes.
Therefore, the most naïve font representation scheme, as a collection of bitmaps for all glyphs in all sizes, is highly infeasible, both in terms of storage space and the amount of work imposed on typeface designers.
A feasible alternative would be to store glyph images only in some key sizes, and use image resizing techniques to obtain the rest of the sizes. However, such a method introduces undesired artifacts, compromising the glyphs' quality (\Cref{resize-issues-med}).

\begin{figure}[t]
  \centering
  \includegraphics[width=.5\linewidth]{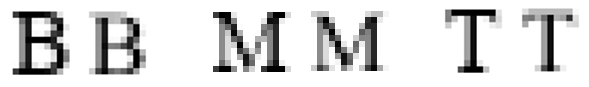}
  \caption{Comparison of vector glyph rasterization with hinting (left) and without hinting (right). The (upscaled) glyphs are Times New Roman B/M/T, pointsize 14. The differences are prominent, e.g.\ note the top of the glyphs, where in the unhinted version the edges are blurred and unclear, as opposed to the hinted version.}
  \label{hinting-importance-med}
\end{figure}

Therefore the prominent representations for glyphs in fonts today use vector graphics. A glyph's outline is defined by a sets of mathematical curves.  Such representation can support multiple sizes simply by converting the vector representaiton to bitmaps in any desired size. Unfortunately, this simple procedure is not sufficient for high-quality glyph rendering, especially for small font sizes, as the rasterization process can also introduce artifacts (\Cref{hinting-importance-med}). To address this, modern font formats include \emph{hints} -- systems of instructions defined by the typeface designer, which are used to modify the glyph outline to better align it with the rasterized grid and avoid undesired artifacts (see \Cref{fig:rasterization}).

\paragraph{Vector Glyph Outlines}
In Vector Graphics, shapes are represented as mathematical objects --- lines, circles, curves, etc.\ in some canonical coordinate system. When an image represented by vector graphics is to be rendered, the mathematical representation is first scaled based on the rendering dimensions, and then converted to a matrix representation via a process called \textit{rasterization}.

\begin{figure}[t]
  \centering
    \hfill
         \includegraphics[width=.3\linewidth]{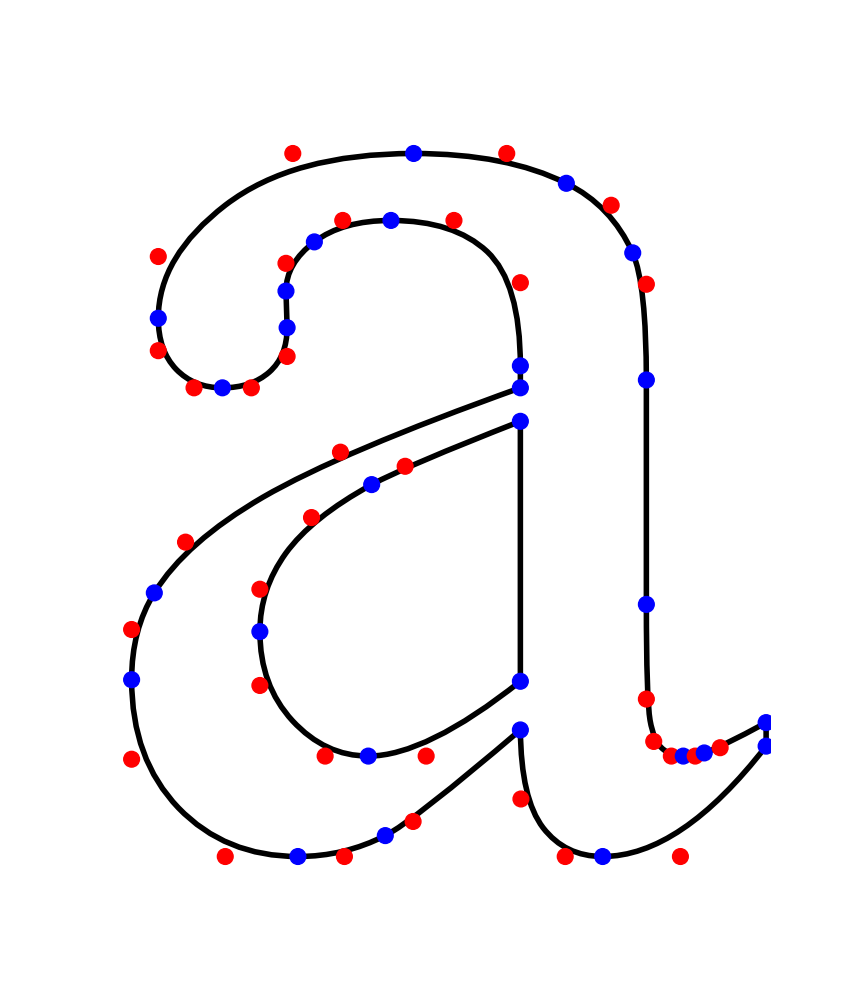} \hfill
         \includegraphics[width=.3\linewidth]{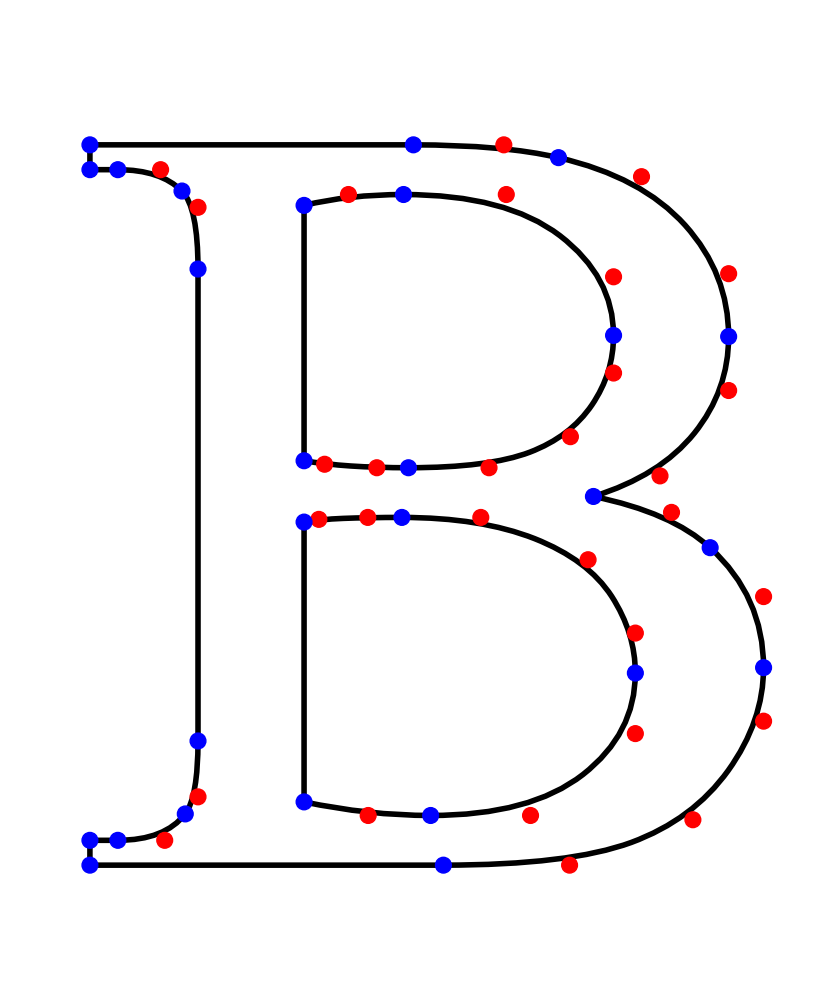} \hfill
         \includegraphics[width=.3\linewidth]{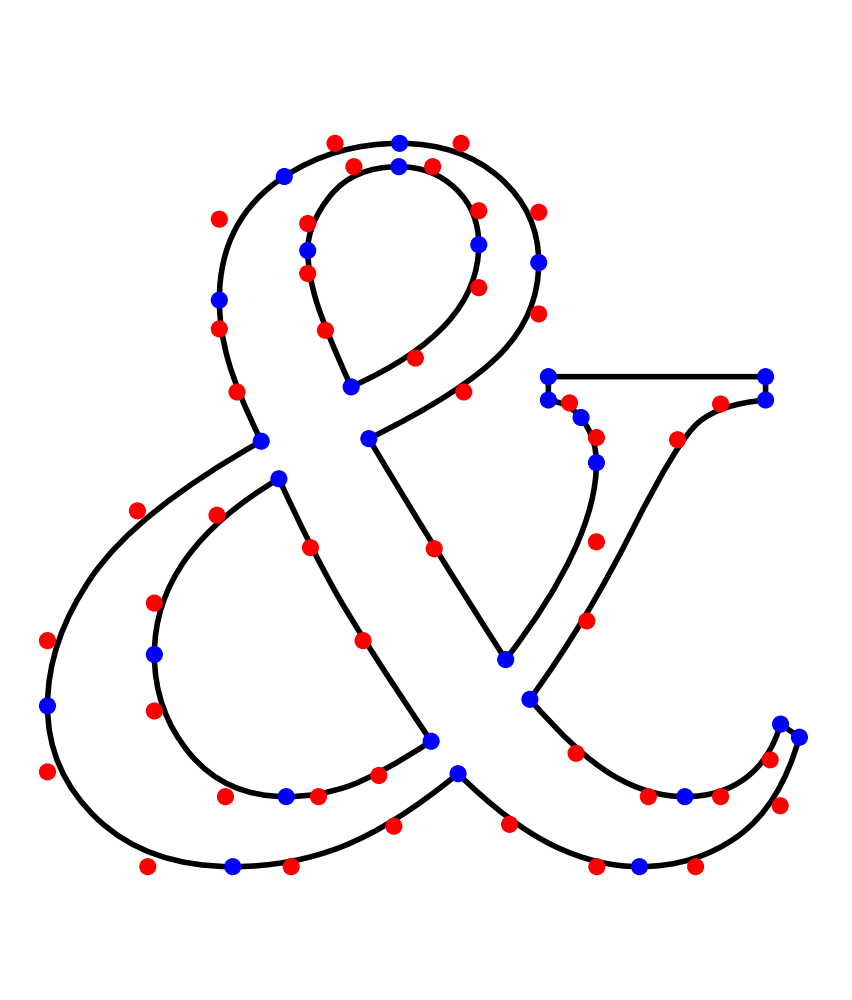} \hfill \hfill
  \caption{Vector graphics representation of the outline of a glyph. The contours are defined by a sequence of points: the blue points lie on the outline (``on points''), while the red points are control points (``off points'') that define e.g.\ a Bézier curve interpolation. }
  \label{vector-representation}
\end{figure}

Modern typeface formats such as TrueType~\cite{TrueTypeHist} and PostScript~\cite{AdobeType1} represent glyphs using a vector graphics representation of their outlines. Specifically, the outline contours are typically represented by collections of lines and Bézier or B-Spline curves. Conveniently, such a representation can be defined by storing only the ordered set of points defining the closed contours, along with tags indicating the types of point (i.e.\ which points are part of the contour and which points are e.g., Bézier control points - see \Cref{vector-representation}).

\paragraph{Hinting}
Simple rasterization is not sufficient to to achieve high quality resulting bitmaps for glyphs. Hence, the hinting process aims to resolve these issues by applying modifications to the glyph outlines, after scaling to a given size but before rasterization. These modifications primarily better align the outline to the rasterized pixel grid. In \Cref{fig:rasterization} the effects of hinting modifications can be seen on the same glyph outline while grid-fitting to two different sizes.

Glyph hints are created as part of the type design process, either using automatic hinting techniques~\cite{Shamir03,Hersch91}, or manually by the typeface designer.
The types of hints, and the ways they are applied, differ between type formats. Typically, hinting systems provide a set of instructions which can be applied to move the contour points depending on the scaling size. These instructions are used to specify typographic constraints or rules, for example, justification of a point based on the global glyph height, or specifying a minimum distance between two points. 
Overall, the resulting glyph rendering process is quite involved, generally consisting of the following steps:
\begin{itemize}
\setlength{\itemsep}{0pt}
\setlength{\parskip}{0pt}
    \item Scaling the vector representation of the glyph outline
    \item Applying hints to the glyph outline
    \item Rasterizing the outline onto the pixel grid
\end{itemize}

A good summary is given by Hersch in~\cite{Hersch94}. Although dated from 1994, this high-level description is still accurate today. We propose an approach for simplifying the glyph rendering process while satisfying all the requirements that led to the complexity of current approach.

\section{Related Work}
We are not aware of related work addressing the main topic of this work, i.e., designing an alternative font representation and rasterization process that does not involve hinting at all (while preserving glyph quality and supporting multi-scale rendering). However, there are several relevant adjacent topics in which related work has been done. In this section we will review these topics and relevant work.

\paragraph{Automatic hint generation}
There has been some work on automating the process of hint specification. While this leaves the rasterization process unchanged, it relieves typeface designers from the need to specify hints, significantly reducing the amount of effort required to design a new typeface.

Hersch and Betrisey~\cite{Hersch91}, create a topological model describing the general shapes of glyphs in Latin typefaces, along with a table of applicable hints for certain sets of characteristic points. Individual glyphs in a new font are then matched to the model, and hints are carried over based on the presence of the required characteristic points. A different approach is taken by Shamir~\cite{Shamir03}, where hints are automatically generated for any typeface (including e.g.\ Chinese and Japanese) in a process of identifying local glyph features that require hinting, collecting global font statistics, prioritizing hints and high-level font consistencies, and converting the results to a hinting specification.

\paragraph{Font synthesis and manipulation}

Considerable work has been done in the area of font generation, completion from several samples, interpolation or other manipulation; this is useful especially in languages with hundreds or thousands of glyphs, such as Chinese.

Suveeranont and Igarashi~\cite{Suveeranont09}, blend a dataset of fonts to generate novel fonts based on several glyph samples. Glyphs are represented by their outlines, and blended using geometric techniques. This yields a vector representation of new fonts. However, hints are not amenable to blending or interpolation, leaving the generated fonts un-hinted and likely to be rasterized with artifacts in small point sizes.

Niell et al.~\cite{Campbell14}, learn a manifold for each glyph. This is done by aligning glyph outlines from different fonts, parameterizing the outlines, and embedding the parameterization in a two-dimensional manifold allowing smooth interpolation between the glyphs of different fonts. A similar approach based on strokes is taken in Balashova et al.~\cite{Balashova2019LearningAS}.
These approaches are limited to outline or stroke representations and do not handle the hinting of the generated glyphs. 

In Phan et al.~\cite{Phan15}, a dataset of fonts is used to learn transfer rules between glyph parts. These can then be used, given several samples of glyphs in a new font, to compose the rest of the glyphs in the font. The glyph part representation consists of strokes and brushes. Another interpolation work is proposed by Bernhardsson~\cite{Bernhardsson2016}, where fixed-size glyph images are parameterized in some latent space using a generative deep learning model. This allows interpolating between glyphs in the latent space and generating the corresponding glyph images. Again, the generated glyphs of these approaches do not handle hints and cannot be scaled to multiple sizes effectively. 

Several other works \cite{Upchurch16,Park2018TypefaceCW,Jiang2019SCFontSC,Xi20} use various deep learning architectures to complete fonts based on one or few glyph samples, often using elements from variational auto-encoders (VAEs) \cite{kingma2014autoencoding} and/or generative adversarial networks (GANs) \cite{goodfellow2014generative}. 
Lian et al.~\cite{Lian16} describe a system for completing Chinese fonts from several sample hand-drawn glyphs. Glyphs are represented as collections of strokes, parameterized by their shape and layout (as deviance from some reference font). For each collection of handwritten glyphs, a neural network is trained to predict the deviance of the strokes from the reference; then the network is applied to predict, for the rest of the glyphs, their shape and layout. In Wang et al.~\cite{Wang20}, fonts are synthesized based on user-provided semantic style attributes. A deep learning model is trained to transform fixed-size glyph images from a source style to a target style. These works represent glyphs as images, and generate the completed font glyphs in a single size. Therefore, to support true scalable font representations, further work is required.

Another line of works \cite{Atarsaikhan17,Azadi2018MulticontentGF,Zhu20,DG-Font} use ideas from neural style transfer \cite{Gatys15}, either directly applying style transfer to stylize glyphs of a certain fonts with samples of a new font, or performing a similar operation with different architectural components. These methods, yet again, operate in a single scale, and require extension to handle multiple sizes.

\paragraph{Memorization in deep learning}
The first part of our work involves training a deep learning model to generate a given font's glyphs in multiple scales, with no generalization requirement, by memorizing the glyphs. Several works have investigated the topic of memorization in deep learning, both as a desired and undesired aspect of learning.

Ulyanov et al.~\cite{Ulyanov17} train a generative network to generate a single image from a random latent vector. The work showed that model architecture could provide a useful prior in the domain of image synthesis, and that the described technique could be used in several image-enhancement tasks such as super-resolution or noise removal.
Arpit et al.~\cite{arpit2017closer} investigate the memorization aspect in the training process of deep learning models, and its relation to other aspects of the model such as generalization and robustness. Among other things, the work highlights that deep learning models, while capable of fitting noise, tend to prioritize learning simple patterns, potentially making near-perfect memorization more difficult to achieve.
In a related vein, Tancik et al.~\cite{tancik2020fourier} showed that in various memorization tasks, deep learning models struggle to memorize high-frequency aspects of the output, unless the input domain is expanded to include frequency information (``Fourier features'').

\paragraph{Implicit Modeling} 
Two other related implicit representations in other domains are SinGAN and NeRF.
SinGAN~\cite{shaham2019singan} is a generative adversarial network~\cite{goodfellow2014generative} that is trained on a single image, enabling generation of similar but not identical images. 
NeRF~\cite{Mildenhall20} is an implicit representation of a textured 3D object that is learned by training a deep learning model to predict pixel values of a rasterization of the object from various angles, allowing novel view synthesis --- images depicting the rendered objects from different angles. These works inspire our implicit representation approach, both conceptually and in architectural terms.

\section{Neural Font Rendering}
We conducted two primary experiments for replacing the traditional font rendering pipeline with a neural-network based approach. In both approaches, we train a model to rasterize a single glyph in \emph{multiple} sizes, optionally allowing specification of font weight. Since glyphs must be rendered in multiple sizes, the network should accordingly support multi-resolution outputs.
This is a fundamental challenge in training neural networks that can be achieved  in several ways:

\begin{itemize}
    \item Use multiple outputs (prediction heads), with every supported size having a dedicated output from the model. However, this approach requires persisting network weights for each bitmap size, in which case it would have been much more memory efficient to simply store the bitmap values themselves for each size.
    \item Use a fixed output size at largest prediction size, and then sample to the desired size. This is the approach employed in our first experiment.
    \item Use an implicit model prediction for a single pixel intensity. In this approach, bitmaps are constructed by sampling the model at multiple pixels. This is the approach employed in our second experiment.
\end{itemize}

\subsection{Masked fixed-size MLP}
\label{sec:maksedMLP}

In our first experiment, we train a model with a fixed-size output to rasterize glyphs in multiple sizes, one model per glyph. This is done by fixing the output size to the maximal desired bitmap size, and sampling it to get smaller bitmap sizes. We use an MLP with a flattened bitmap output, whose size is determined by the largest supported rendering size. The desired point size is indicated as input to the model (in a one-hot encoded fashion). 

\paragraph{Sampling Representation}
A key aspect in this approach is the sampling strategy; for example, a simple sampling strategy is to crop the output to the desired bitmap size, for instance, around the center of the bitmap. Such an approach suffers from a drawback, in that different point sizes require very different (non-smooth) predictions from the model. Since there are multiple ways to embed smaller sizes in the full bitmap, we choose a configuration that aligns the relative position of pixels as much as possible, so as to minimize the variation in ground truth intensity of each pixel as a function of the point size. 

\begin{figure}[t]
  \centering
  \includegraphics[width=\linewidth]{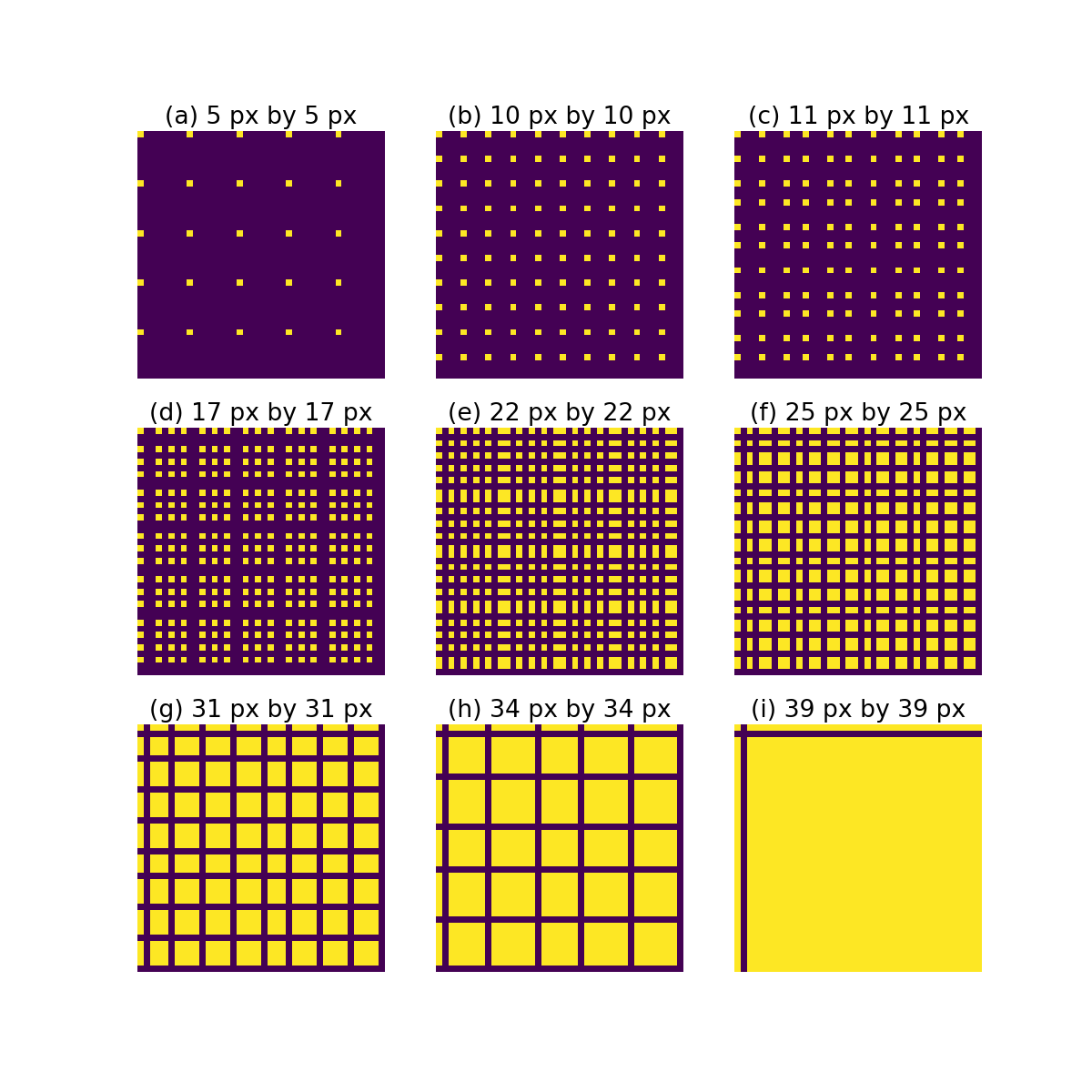}
  \caption{Masks indicating subsampling strategies for a renderning size of 40 pixels by 40 pixels, for various original image sizes. Purple color indicates 0 in the mask, and yellow color indicates 1.}
  \label{subsampling-strategy}
\end{figure}

Our sampling approach is motivated by upsampling methodology. 
Assume you are given a glyph bitmap and you need to upsample it to a larger target size (in practice, we assume both dimension are equal). Essentially, we want to upsample the bitmap to the desired size in a manner that preserves the relative position of each pixel as much as possible. To do this, we first duplicate all pixels in both dimensions an equal number of times as much as possible without exceeding the target size. Then, we uniformly choose additional pixels to duplicate in order to match the target size exactly. By tracking the positions of the original pixels of the bitmap, we can define a mask that can be used to sample the larger size image to get back the smaller size bitmap. Since this can be done for any size, we fix the target large size and define a mask for any smaller-size bitmap using this strategy (see \Cref{subsampling-strategy}).

\begin{figure}[t]
  \centering
  \includegraphics[width=\linewidth]{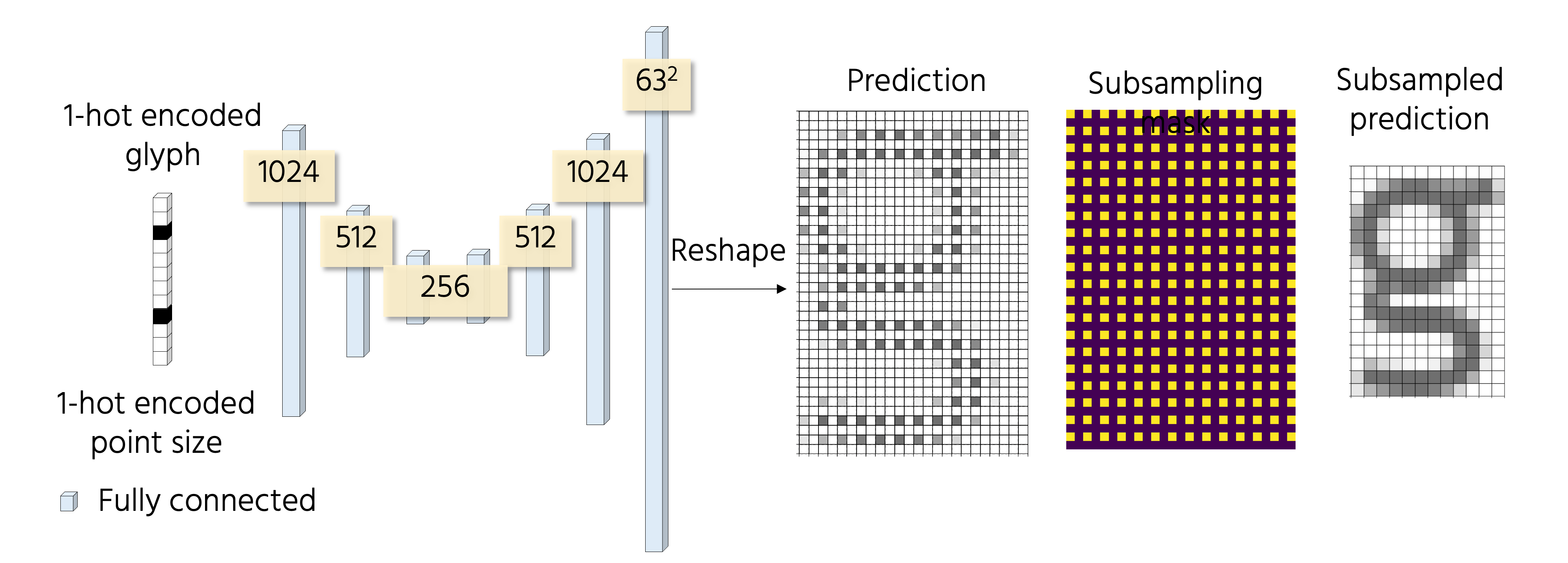}
  \caption{The architecture of the Masked-MLP of our first experiment is basically a U-net encoder-decoder trained on all sizes of a given glyph. }
  \label{fig:arch-3}
\end{figure}

\paragraph{Training}
During training, the pixel loss is calculated using the upscaled versions of the glyphs, masked according to the sampling strategy. This means only pixels in the actual bitmaps are accounted for in the loss - focusing training on the pixels relevant to the bitmap being rendered. Such a loss is equivalent to first sampling and then calculating the loss on the glyphs in their target size, except it is more efficient in terms of vector operations.  The loss function is L2 pixelwise loss. The architecture is depicted in \Cref{fig:arch-3}.

In experiments testing this approach, the model was trained for 1600 epochs using the Adam optimizer \cite{Kingma2015AdamAM} with the learning rate alternating between 0.0005 and 0.00001 every 50 epochs.

\subsection{Implicit Representation}
\label{sec:implicitRep}

In the previously presented approach, the rendered bitmap was directly obtained as the network's output. However, there are other approaches to image generation, which do not immediately yield the full image, but rather create it by sampling each pixel. A recent example is NeRF~\cite{Mildenhall20}, in which a model is used to obtain 3D scene density and color values for given positions viewed from given angles.  Inspired by this work, we use a rasterization process based on sampling: we train a model to predict the intensity of a single pixel, conditioned both on the the pixel position and the glyph parameters (size, weight). The full glyph image is composed by sampling the bitmap at all the necessary pixels. Such an approach naturally allows rasterization of glyphs in multiple sizes.

\begin{figure}[t]
  \centering
  \includegraphics[width=\linewidth]{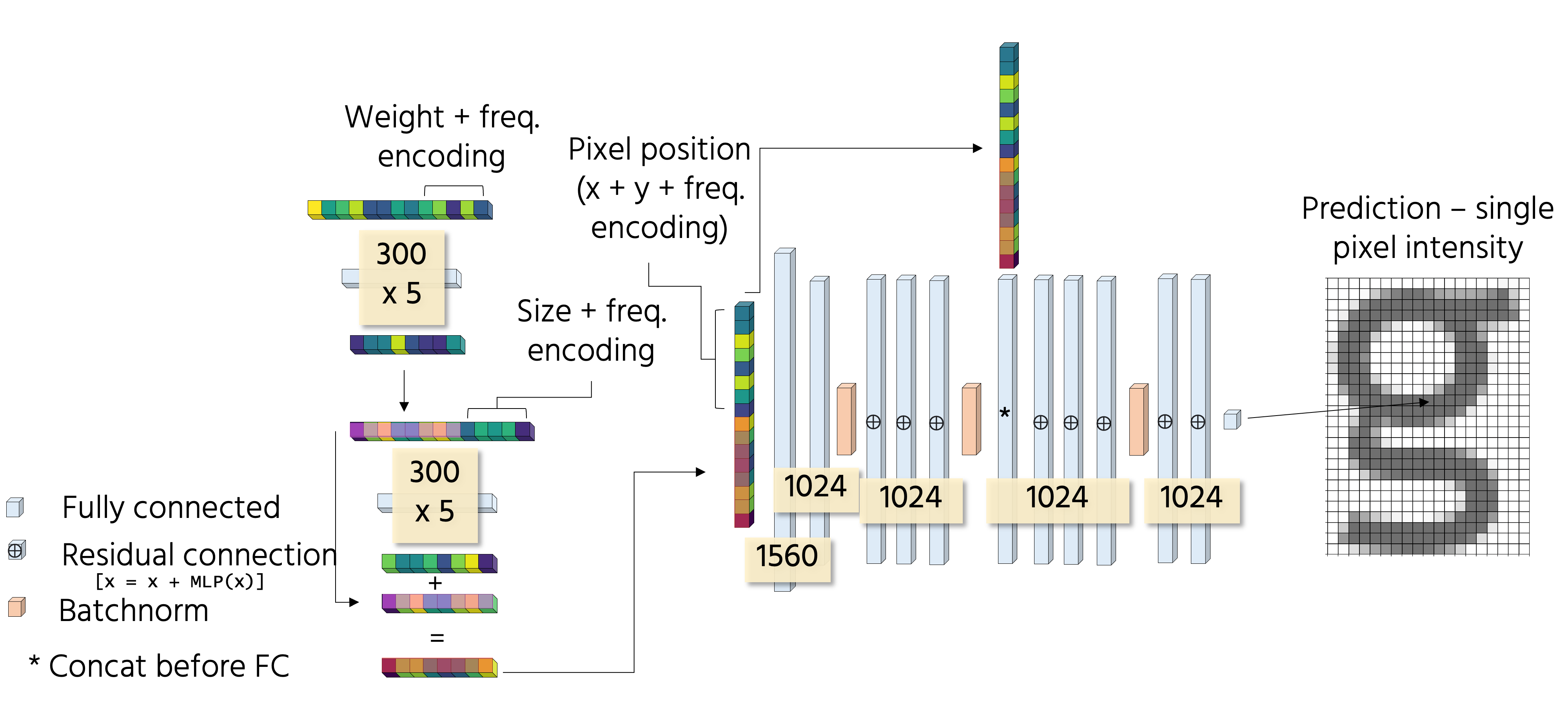}
  \caption{ The architecture of the Implicit model of our second experiment, consisting of a deep representation of the shape, and an MLP predicting the intensity of the shape in a given (x, y) position (with positional encodings). Each deep representation MLP consists of 5 layers, and the intensity prediction MLP consists of 11 layers. }
  \label{fig:arch-4}
\end{figure}

The model's architecture is described in \Cref{fig:arch-4}. We use the model to represent a function, mapping pixel positions to pixel intensities, thereby enabling sampling-based rendering in multiple sizes. The input to the model consists of:
\begin{itemize}
    \item Glyph size (along with frequency encoding).
    \item $(x, y)$ position of the pixel to be sampled, along with frequency encoding of the position (see explanation below). 
    \item (Optionally) glyph weight.
\end{itemize}

The model outputs a number between 0 and 1, indicating the sampled pixel's intensity.

\paragraph{Frequency encoding} 
We use frequency encoding similar to the positional encoding of~\cite{Mildenhall20}, for the same reason - allowing the model to more easily capture high-frequency information. Each encoded scalar $v$ is projected to the $2D$-dimensional vector $\gamma(v)$, defined as:

$\gamma(v)_{2i} = \sin(2^{L_i}\cdot \pi \cdot v)$

$\gamma(v)_{2i + 1} = \cos(2^{L_i}\cdot \pi \cdot v)$

Where $L$ is a list of powers. In our experiments, we used $D=32$ with $L$ sampled uniformly in the range $[0, 12]$.

\paragraph{Training}
Our experiments differ from more traditional applications of deep-learning models in a subtle but important manner. Usually, models are expected to be used on new, unseen data; therefore, some error in performance is tolerated (sometimes even desired to avoid overfitting), as long as the model generalizes well to unseen data. In our case, however, we are interested in having the model memorize the training data with as little error as possible. To that end, several small modifications proved crucial, related to the batch normalization layers \cite{pmlr-v37-ioffe15}:
\begin{itemize}
    \item During training, the batch normalization layer normalizes by magnitudes observed in the current batch, while collecting normalization statistics. During inference, the normalization is performed using statistics collected during training, leading to slightly different behaviour compared to the training phase. In our experiments, we adapted the model to always behave as in the training phase so that inference would be identical to training.
    \item Since the normalization depends on batch magnitudes, the traditional practice of shuffling batches leads to stochastic behaviour. To avoid this stochasticity, we train with a batch size of 1, i.e.\ each glyph is always processed individually by the model.
\end{itemize}

These adaptations had a significant effect on the memorization quality, as will be demonstrated in the ablation section. The target pixel intensity was quantized to 20 categories, and the model was trained using focal loss \cite{focal_loss} on predicting the correct intensity category.
In experiments testing this approach, the model was trained for 500 epochs using the Adam optimizer \cite{Kingma2015AdamAM} with an exponentially decaying learning rate that is reset every 100 epochs, according to a learning rate schedule (refer to the supplemental materials for the detailed schedule).

\subsubsection{Interpolation} 
\label{interpolation}
When training a glyph model on more than a single font, interpolation between styles can be performed. For example, we can train a model to rasterize a glyph in multiple weights of the same font. In this case, we can attempt to rasterize the glyph in a novel weight between existing weights. To do this, we use linear interpolation in the latent representation space. We use the network vector layer before the pixel intensity prediction MLP along with pixel position.

\section{Experiments and results}
\begin{figure*}[t]
  \centering
  \includegraphics[width=0.95\linewidth]{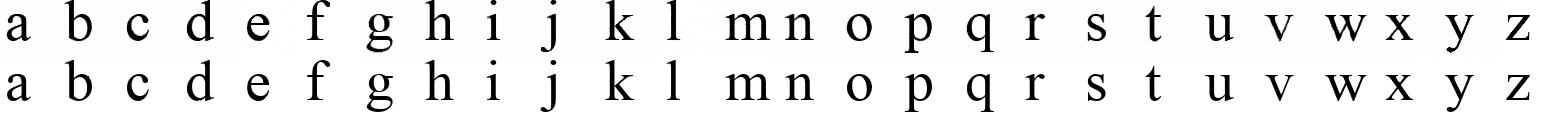}
  \caption{ Masked MLP (top) and Implicit representation (bottom) predictions for all lowercase glyphs, for Times New Roman, bitmap size 60. More results can be seen in \Cref{full_result_visualization}. }
  \label{fig:times-new-roman-small-cascade}
\end{figure*}

We performed several experiments with the apprpoaches described in the previous section. All experiments involved training models to rasterize glyphs in bitmap sizes between 20 and 63 (inclusive), where point size for the ground truth for a certain bitmap size was determined by finding the largest point size in which all rasterized glyphs fit in the bitmap.
\Cref{fig:times-new-roman-small-cascade} demonstrates sample results for both approaches. For additional results (more sizes and fonts), please refer to \Cref{full_result_visualization}.

\subsection{Ablations}

\begin{figure}[t]
  \centering
  \includegraphics[width=\linewidth]{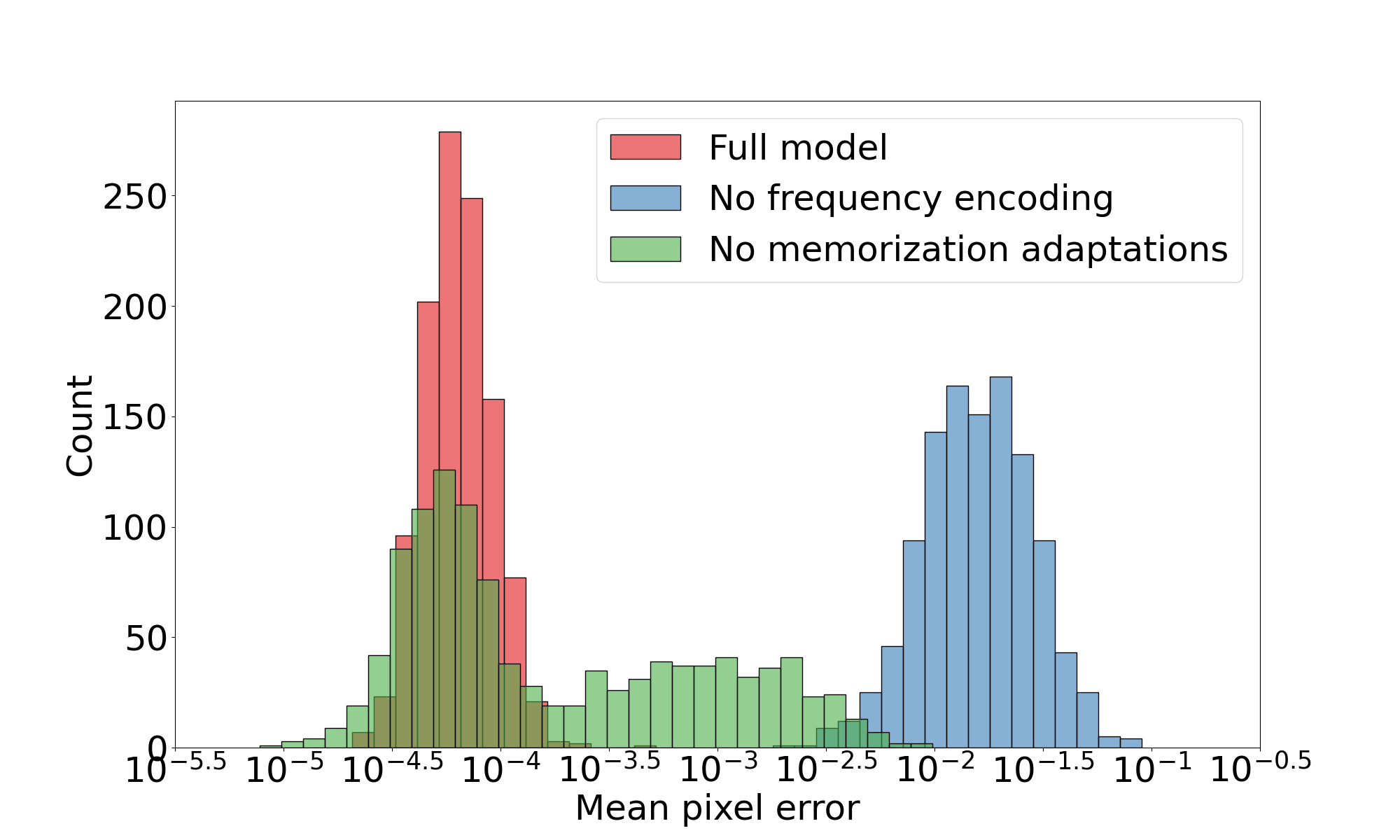}
  \caption{ Ablation results - comparing of the histograms of errors in two ablation experiments vs. full training regime. It is evident that partial training regimes (no frequency encoding / no memorization adaptations) result in larger errors. }
  \label{fig:ablation-comparison}
\end{figure}
\begin{figure}[t]
    \centering
    \begin{subfigure}[t]{0.49\linewidth}
         \centering
         \includegraphics[width=0.49\linewidth]{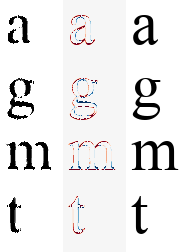}
         \caption{Training without frequency encoding.}
     \end{subfigure}
     \hfill
     \begin{subfigure}[t]{0.49\linewidth}
         \centering
         \includegraphics[width=0.49\linewidth]{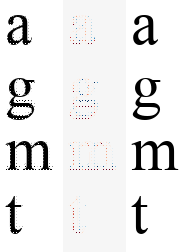}
         \caption{Training without memorization adaptations.}
     \end{subfigure}
     
    \caption{Visualization of ablation results for Times New Roman at bitmap size 63. In both cases the results are visually worse than the full model. Left column shows the predicted bitmap; right column shows the ground truth; and center column shows the error.}
    \label{fig:ablation_viz}
\end{figure}

We performed ablation experiments to demonstrate the importance of two key characteristics of the approach - frequency encoding and memorization adaptations. \Cref{fig:ablation-comparison} shows the experiment results, showing larger error for the partial experiments compared to the full training regime. \Cref{fig:ablation_viz} shows sample results, which visually corroborate the poor performance of these experiments.

\subsection{Approach comparison}
We compare the performance of the two approaches (masked MLP and implicit representation) across three fonts (Arial, Tahoma, and Times new Roman), for the 26 lowercase letters in the English alphabet (a-z).

\begin{figure}[t]
    \centering
    \includegraphics[width=\linewidth]{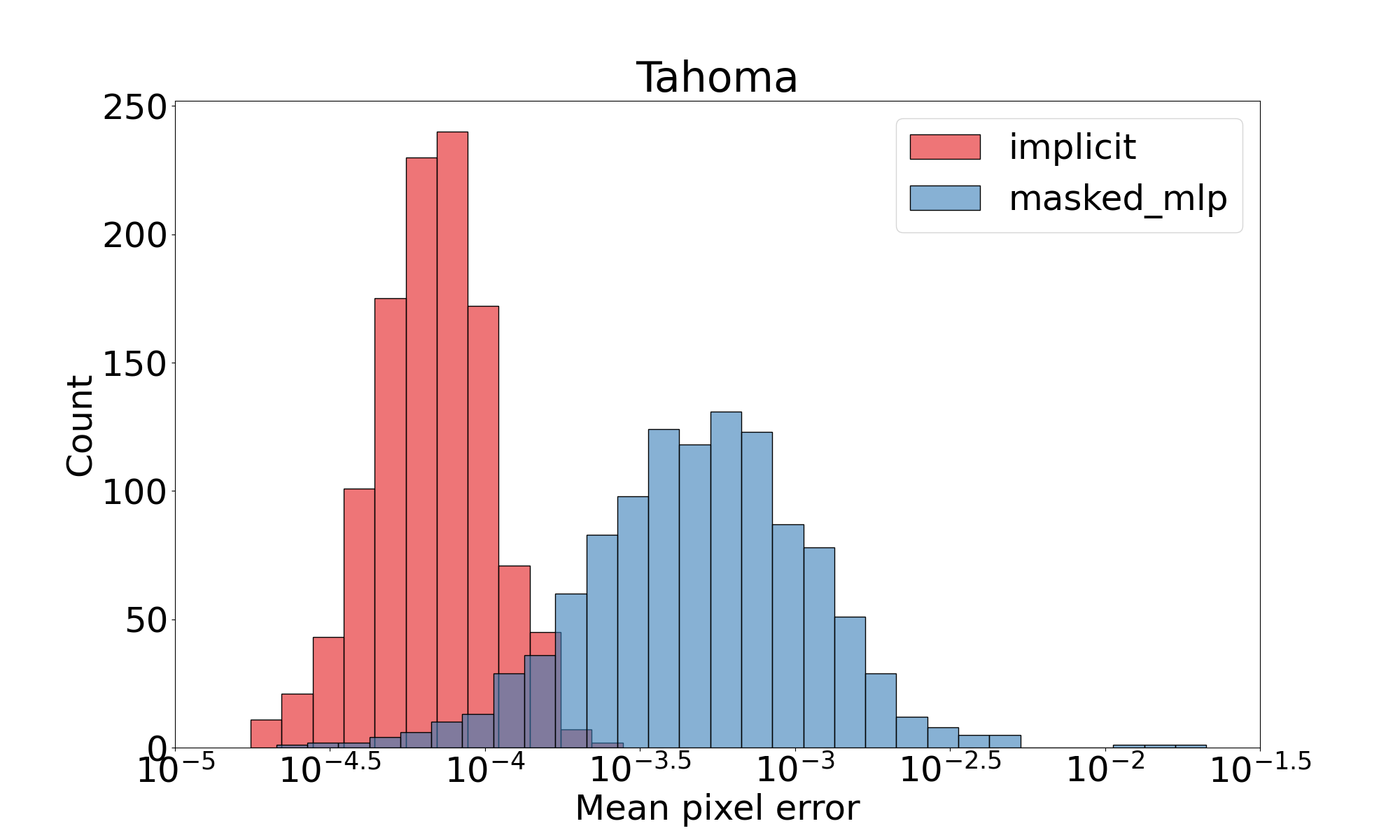}
    \caption{Comparisons of the error histograms between the two approaches we investigated - masked MLP and implicit representation, for the Tahoma font alphabet. It is clear that the implicit representation approach achieves much smaller errors than the masked MLP approach. Other fonts exhibit similar performance behaviors. }
    \label{fig:comparison}
\end{figure}

\Cref{fig:comparison} compares the errors achieved by the two approaches; it is evident that the implicit representation approach was superior to the masked MLP approach by more than an order of magnitude. For comparison of experiments on additional fonts, please refer to \Cref{approach_comparison_across_fonts}.

\subsection{Implicit error analysis}
We performed further analyses on the implicit representation results, visualizing typical error by size and by glyph.

\paragraph{By size}

\begin{figure}[t]
  \centering
  \includegraphics[width=\linewidth]{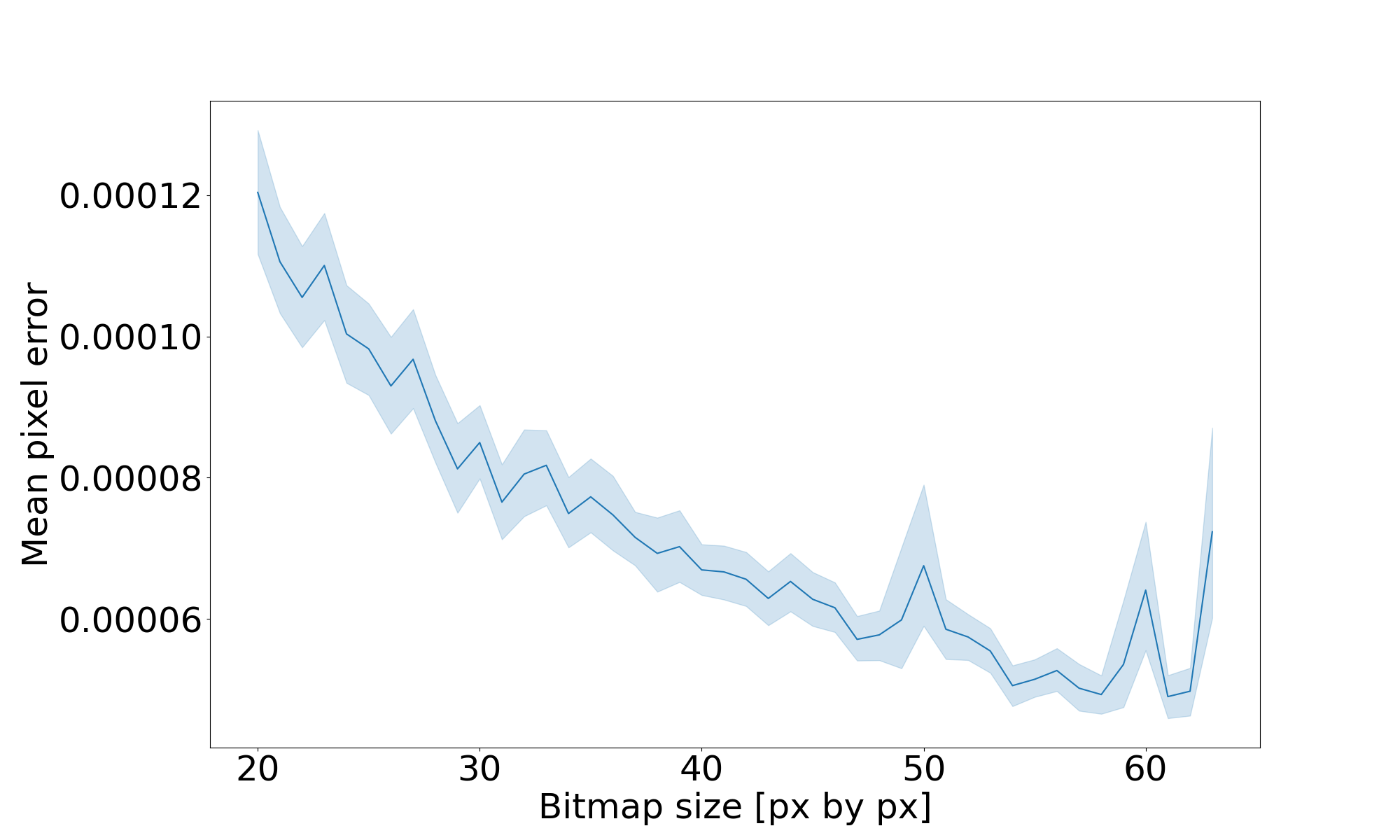}
  \caption{ The mean error (and stdv) of the Implicit representation per bitmap size. Generally, as the bitmap size increases, the mean pixel error decreases. }
  \label{fig:size-comparison}
\end{figure}

\Cref{fig:size-comparison} depicts the typical mean pixel error by bitmap size. The mean error for each size considers all lowercase glyphs for three fonts: Times New Roman, Arial and Tahoma. A general trend can be seen of the mean pixel error decreasing as the bitmap size increases. We hypothesise that this is due to the fact that glyph bitmaps exhibit sharp edges which are challenging for the model to learn, and that the trend can be seen because in smaller bitmaps the transition pixels constitute a larger relative part of the bitmap. 

\paragraph{By glyph}

\begin{figure}[t]
  \centering
  \includegraphics[width=\linewidth]{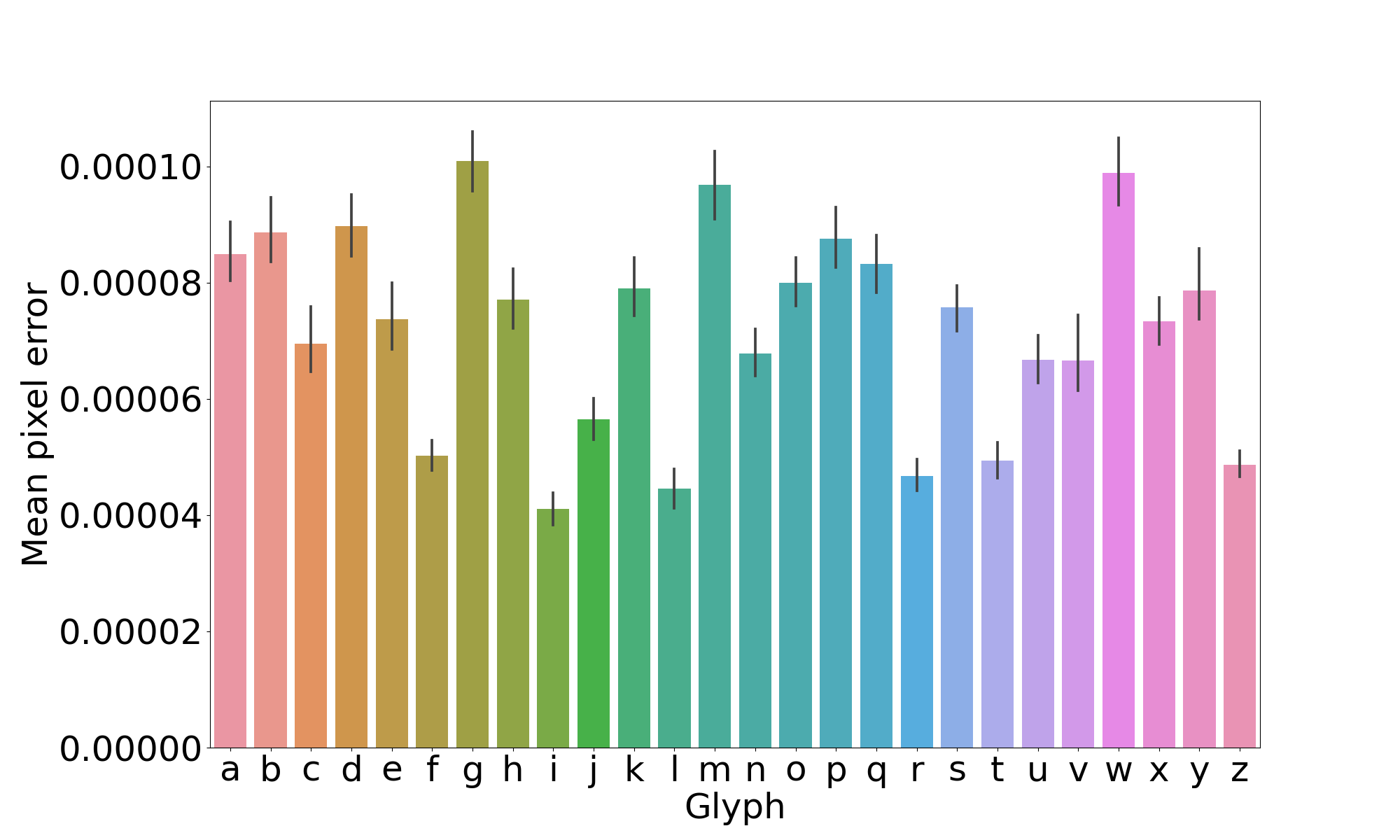}
  \caption{ The mean pixel error by glyph of the implicit representation. }
  \label{fig:glyph-comparison}
\end{figure}
\Cref{fig:glyph-comparison} depicts the typical mean pixel error by glyph. Significant differences can be seen between the glyphs, raising questions about glyph characteristics that make them easier / more difficult for a model to learn. Not surprisingly, simple glyphs, composed primarily of a single line, such as "l" and "i", seem easiest to learn. More complex glyphs, consisting of multiple lines and/or round curves, such as "g", "m", and "w" seem most difficult.

\subsection{Weight interpolation}

\begin{figure}[t]
  \centering
  \includegraphics[width=\linewidth]{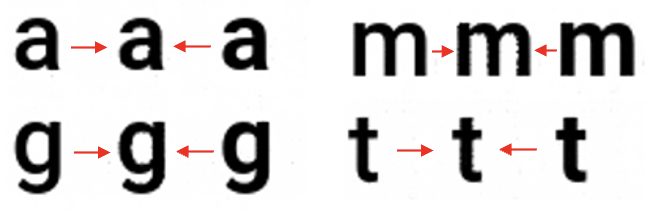}
  \caption{ Weight interpolation results: Ground truth Medium (left) and Bold (right) weights, and predicted Semi-bold weight (center). }
  \label{fig:weight-interpolation-results}
\end{figure}

In this experiment we perform weight interpolation as described in \Cref{interpolation}. We show examples on four glyphs ('a', 'g', 'm', 't'), where we train a model to render multiple weights of Roboto, and perform interpolation to obtain intermediate weights not present in the training set. Specifically, the weights Medium and Bold are interpolated to obtain the Semi-bold weight. \Cref{fig:weight-interpolation-results} presents the interpolation results. These results go a step towards generalization, but as can be seen, still contain artifacts.

\section{Discussion}

\paragraph{Limitations and future work}
There are several aspects in which we think further improvements can be made to the implicit representation approach:
\begin{itemize}
    \item Scale generalization: while we trained the models on all sizes within the range, it could be possible to train only on some sizes, and incorporate in the model an inductive bias that yields good scale generalization. This will enable creation of fonts using less ground truth data, making it feasible for use in typography pipelines.
    \item Other generalizations: weight interpolation still needs fine-tuning to achieve high quality results. It can also be interesting to investigate other dimensions such as style -- it could be possible to train a single model to rasterize the same glyph but from multiple typefaces. This, in turn, might allow interpolating between styles and generating novel fonts as a mixture of existing fonts.
    \item Model size compression: the current architecture results in relatively large models (5M parameters, 24MB on disk). Architecture improvements can allow decreasing this size, or jointly training a single model to rasterize multiple glyphs (especially if they have similar topology or similar elements like `b', `d', `p' and `q'), reducing the overall memory required for a full font.
\end{itemize}

\paragraph{Conclusions}
Based on the results, we believe that our second approach (implicit representation) constitutes a viable first step on the path to representing a scalable fonts with deep learning models replacing the current outline-hinting-rasterization scheme. Our experiments show an architecture that:

\begin{itemize}
    \item Is consistently capable of producing all glyphs in full continuous scales of bitmaps.
    \item Can readily be expanded to manipulate the rasterization process in flexible ways, demonstrated by training a model to rasterize not only multiple scales but also multiple weights, and enabling weight interpolation.
\end{itemize}

Our hope is that this research will promote others to benchmark font representations and stylization not only in large sizes, or in a vector representation and address the challenge of rasterization in a range of continuous sizes similar to true font representations today.

{\small
\bibliographystyle{ieeefullname}
\bibliography{references}
}

\appendix

\section{Approach performance comparison across fonts}
\label{approach_comparison_across_fonts}

\begin{figure}[t]
    \centering
    \begin{subfigure}[b]{0.3\textwidth}
         \centering
         \includegraphics[width=\textwidth]{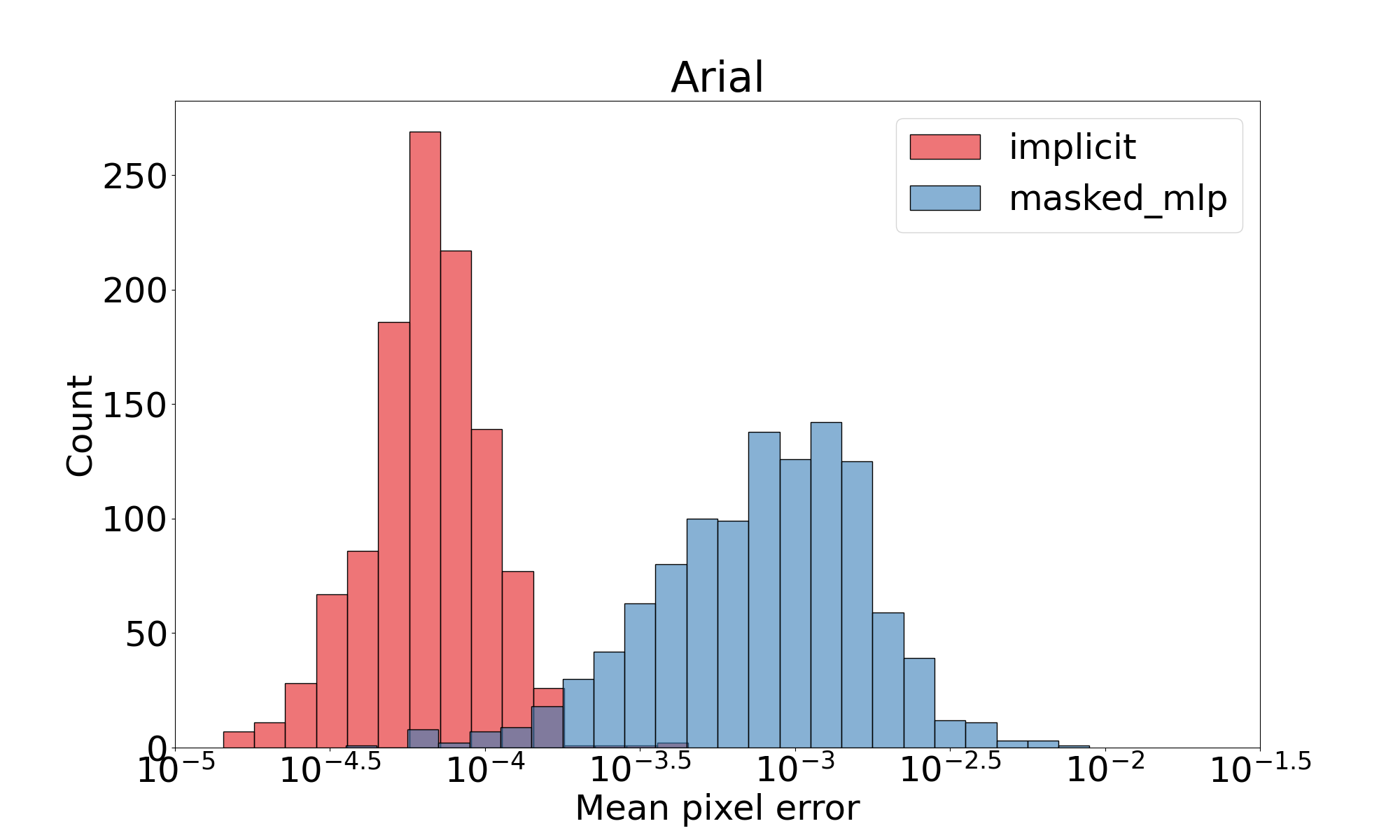}
         \caption{Arial performance comparison}
         \label{fig:arial_comparison}
     \end{subfigure}
     \hfill
     \begin{subfigure}[b]{0.3\textwidth}
         \centering
         \includegraphics[width=\textwidth]{images/tahoma_comparison}
         \caption{Tahoma performance comparison}
         \label{fig:tahoma_comparison}
     \end{subfigure}
     \hfill
     \begin{subfigure}[b]{0.3\textwidth}
         \centering
         \includegraphics[width=\textwidth]{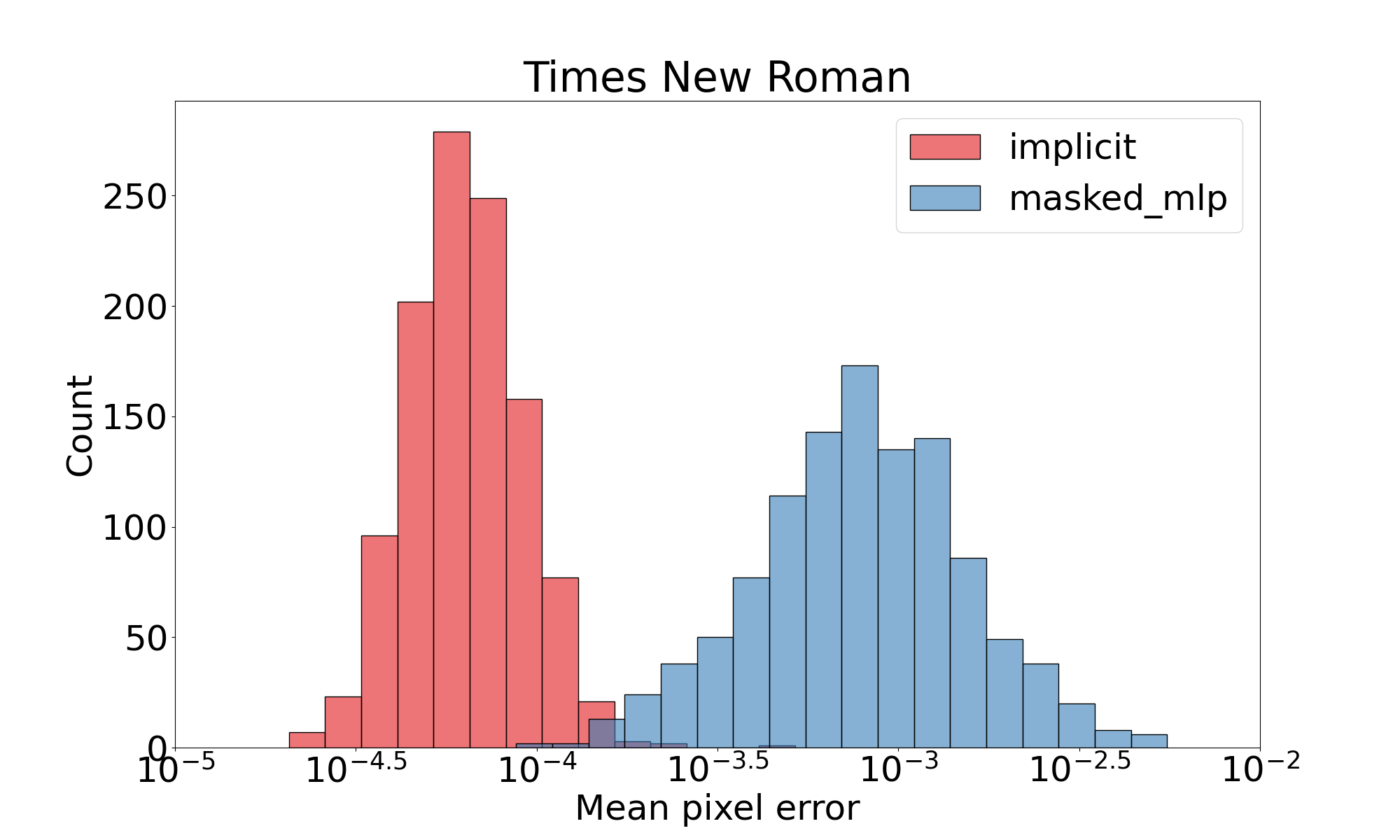}
         \caption{Times New Roman performance comparison}
         \label{fig:times_new_roman_comparison}
     \end{subfigure}
    \caption{Error comparison between the two approaches --- masked MLP and implicit representation. It is clear that the implicit representation approach achieves much smaller errors than the masked MLP approach.}
    \label{fig:comparison_full}
\end{figure}

\Cref{fig:comparison_full} compares the performance of the two approaches for all 3 fonts in the experiment (Times New Roman, Arial and Tahoma). Similar results can be observed --- significantly better performance for the implicit approach.

\section{Implicit approach learning rate schedule}
The following table describes the learning rate schedule for the implicit approach experiment:

\begin{center}
\begin{tabular}{c c c} 
 \toprule
 epochs & initial LR & final LR \\ %
 \midrule
 1--100 & $1e^{-3}$ & $1e^{-5}$ \\ 
 101--200 & $1e^{-4}$ & $1e^{-6}$ \\ 
 201--300 & $1e^{-4} / 2$ & $1e^{-7}$ \\ 
 301--400 & $1e^{-5}$ & $1e^{-8}$ \\ 
 401--500 & $1e^{-6}$ & $1e^{-9}$ \\ 
 \bottomrule
 
\end{tabular}
\end{center}

\begin{figure}[t]
    \centering
    \includegraphics[width=\linewidth]{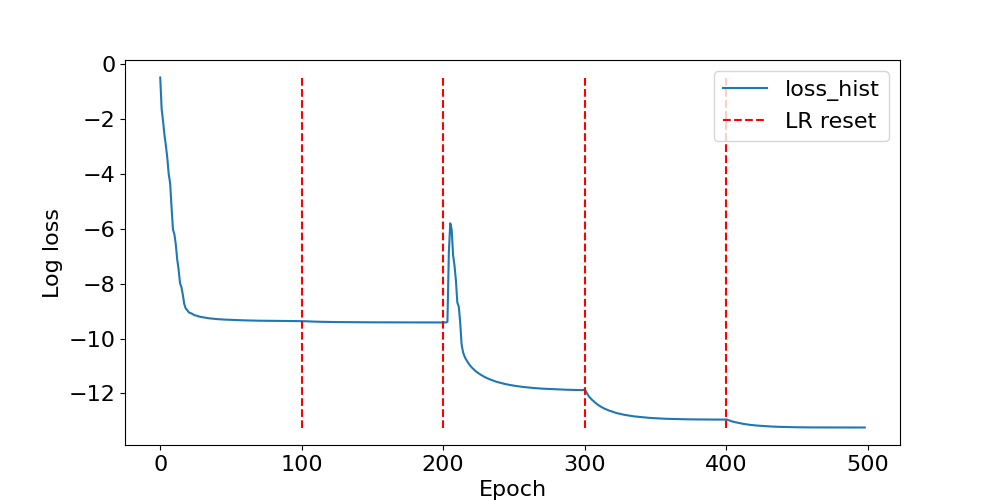}
    \caption{Log loss history of implicit representation model training on lowercase ``m'', Times New Roman, with the learning rate reset points visualized as well. The effect of resetting the LR is evident, allowing the optimization to escape from local optima. Other experiments exhibit similar behavior. }
    \label{fig:log_loss_hist}
\end{figure}

\Cref{fig:log_loss_hist} shows the log loss training history of a sample model, demonstrating how the LR schedule affects the training process and allows the model to escape local optima.

\section{Full result visualization}
\label{full_result_visualization}

\begin{figure*}[hbtp]
  \centering
  \includegraphics[width=.9\textwidth]{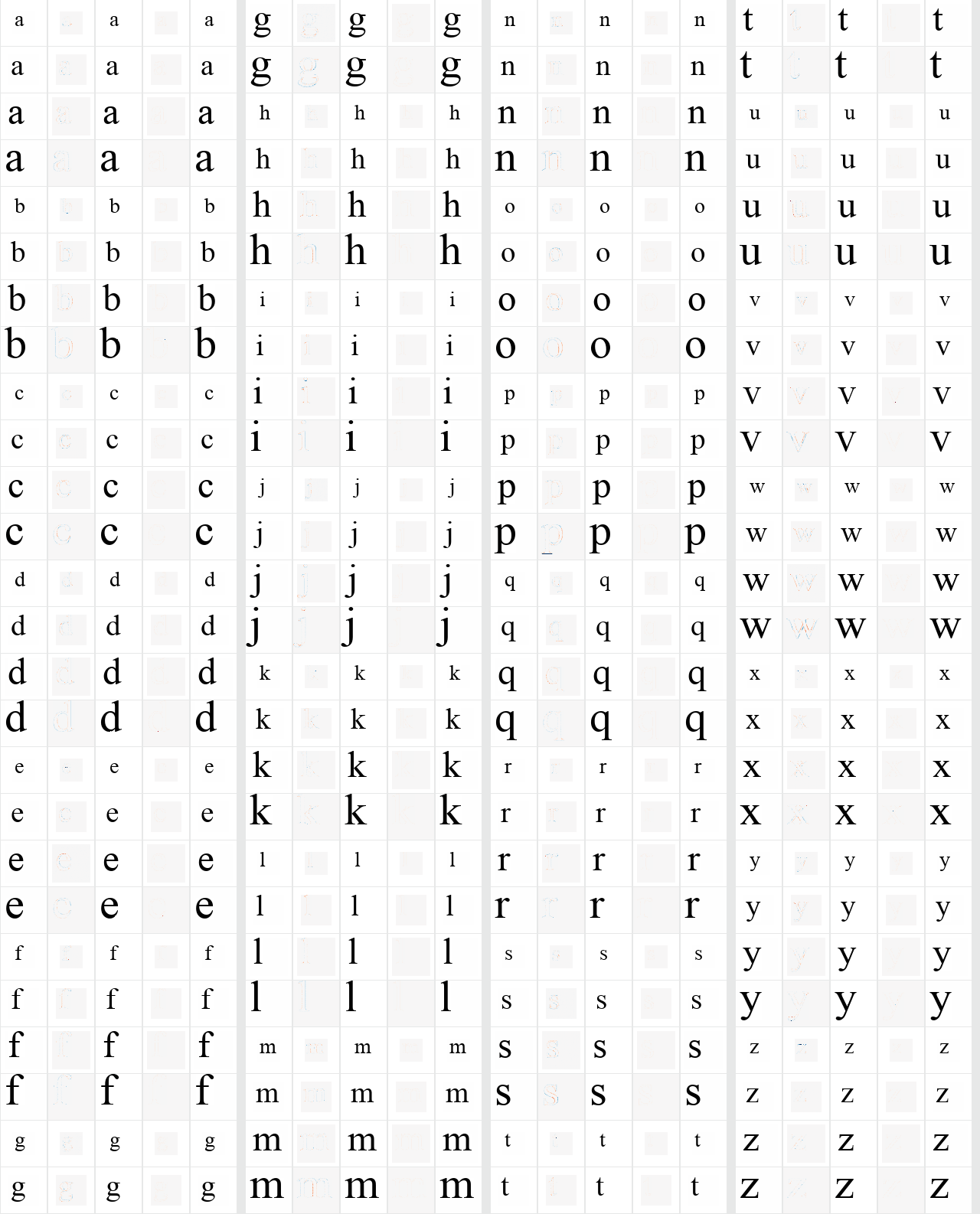}
  \caption{ Times New Roman experiment results. All lowercase glyphs are displayed at four bitmap sizes (30px, 40px, 50px, 60px). The leftmost column displays the masked MLP prediction; the rightmost column displays the implicit approach prediction; the center column displays the ground truth; and the second and fourth columns display, respectively, the masked MLP and implicit approach errors. It is easy to see the better performance of the implicit representation approach, especially in the error visualizations.}
  \label{fig:full-results-times-new-roman}
\end{figure*}

\begin{figure*}[hbtp]
  \centering
  \includegraphics[width=.9\textwidth]{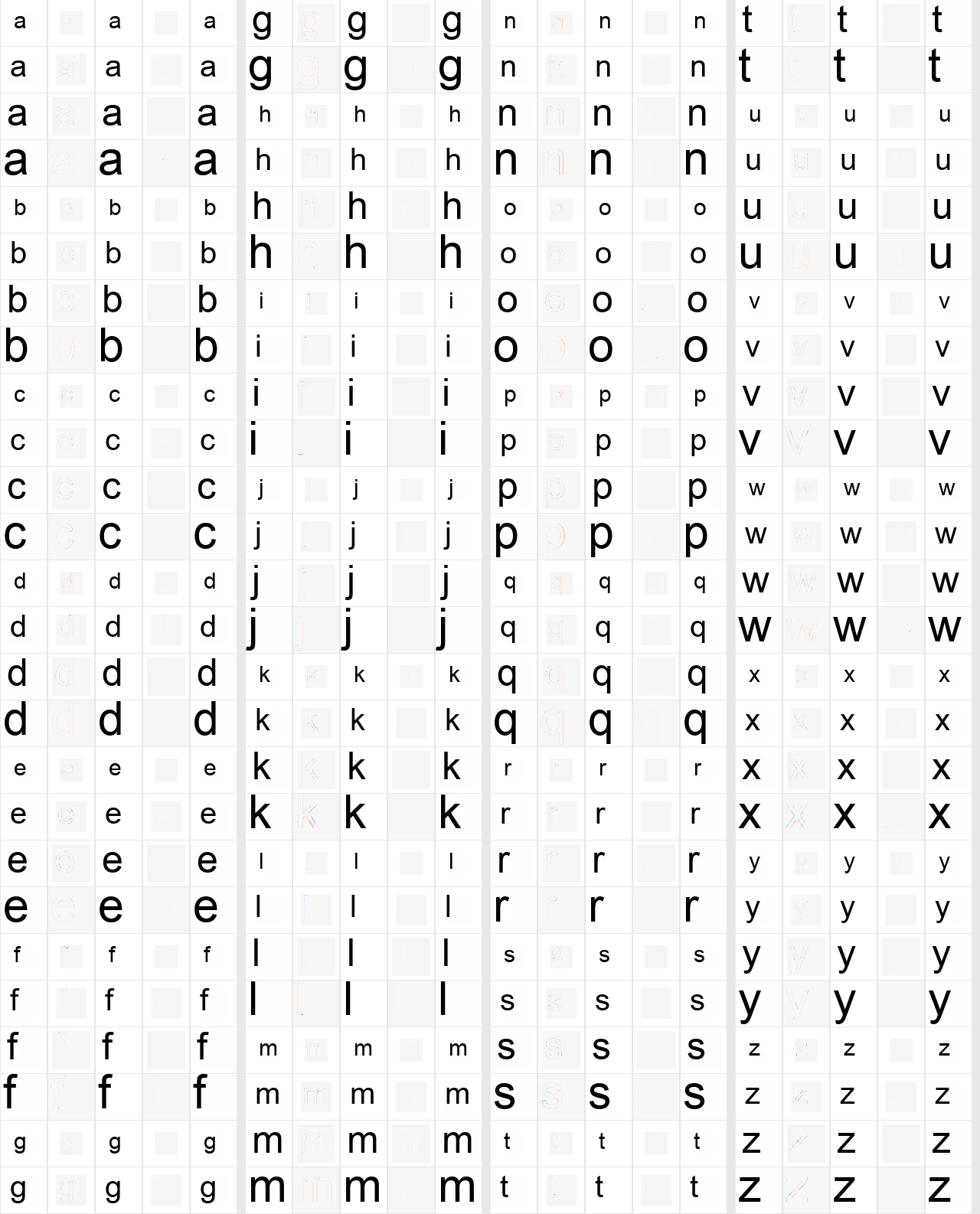}
  \caption{ Arial experiment results. All lowercase glyphs are displayed at four bitmap sizes (30px, 40px, 50px, 60px). The leftmost column displays the masked MLP prediction; the rightmost column displays the implicit approach prediction; the center column displays the ground truth; and the second and fourth columns display, respectively, the masked MLP and implicit approach errors. It is easy to see the better performance of the implicit representation approach, especially in the error visualizations. }
  \label{fig:full-results-arial}
\end{figure*}

\begin{figure*}[hbtp]
  \centering
  \includegraphics[width=.9\textwidth]{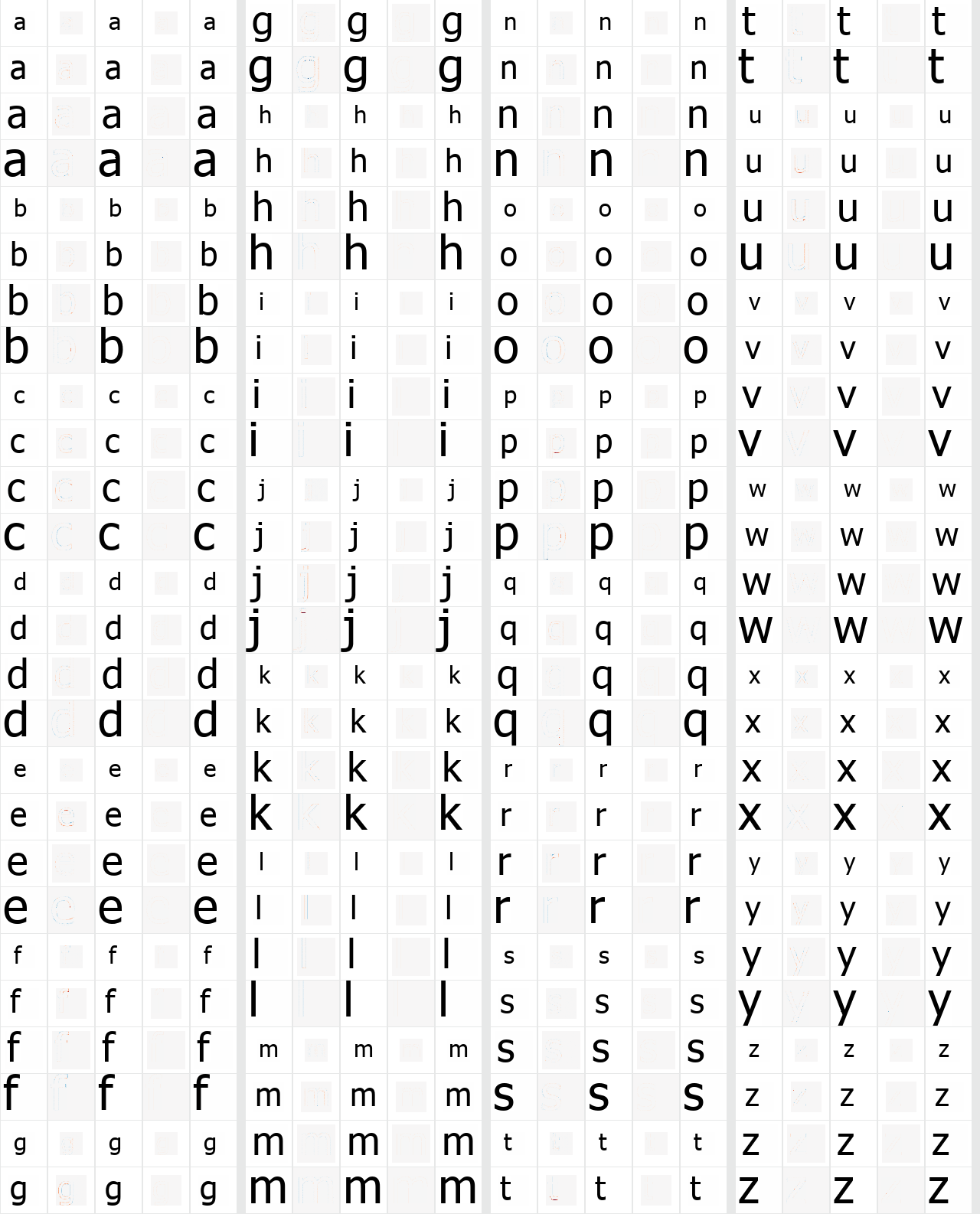}
  \caption{ Tahoma experiment results. All lowercase glyphs are displayed at four bitmap sizes (30px, 40px, 50px, 60px). The leftmost column displays the masked MLP prediction; the rightmost column displays the implicit approach prediction; the center column displays the ground truth; and the second and fourth columns display, respectively, the masked MLP and implicit approach errors. It is easy to see the better performance of the implicit representation approach, especially in the error visualizations.}
  \label{fig:full-results-tahoma}
\end{figure*}

\Cref{fig:full-results-times-new-roman,fig:full-results-arial,fig:full-results-tahoma} demonstrate the results for both approaches on all lowercase glyphs for Times New Roman, Arial, and Tahoma, respectively. The residuals are visualized as well, and the better performance of the implicit representation approach can easily be seen.

\begin{figure*}[hbtp]
  \centering
  \includegraphics[width=.7\textwidth]{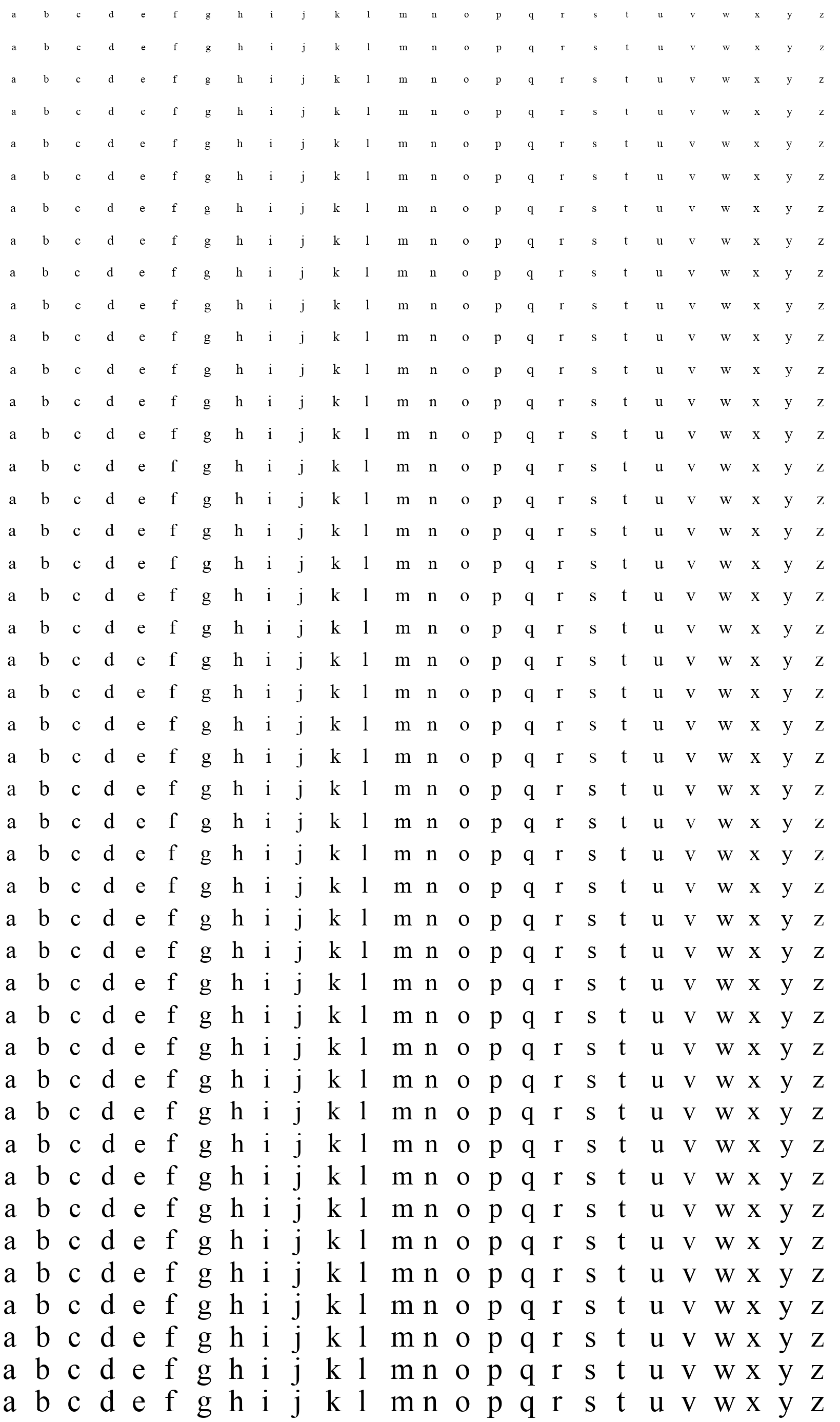}
  \caption{ Times New Roman implicit representation experiment results. All lowercase glyphs are displayed at multiple bitmap sizes (20px - 63px).}
  \label{fig:full-cascade-times-new-roman}
\end{figure*}

\begin{figure*}[hbtp]
  \centering
  \includegraphics[width=.7\textwidth]{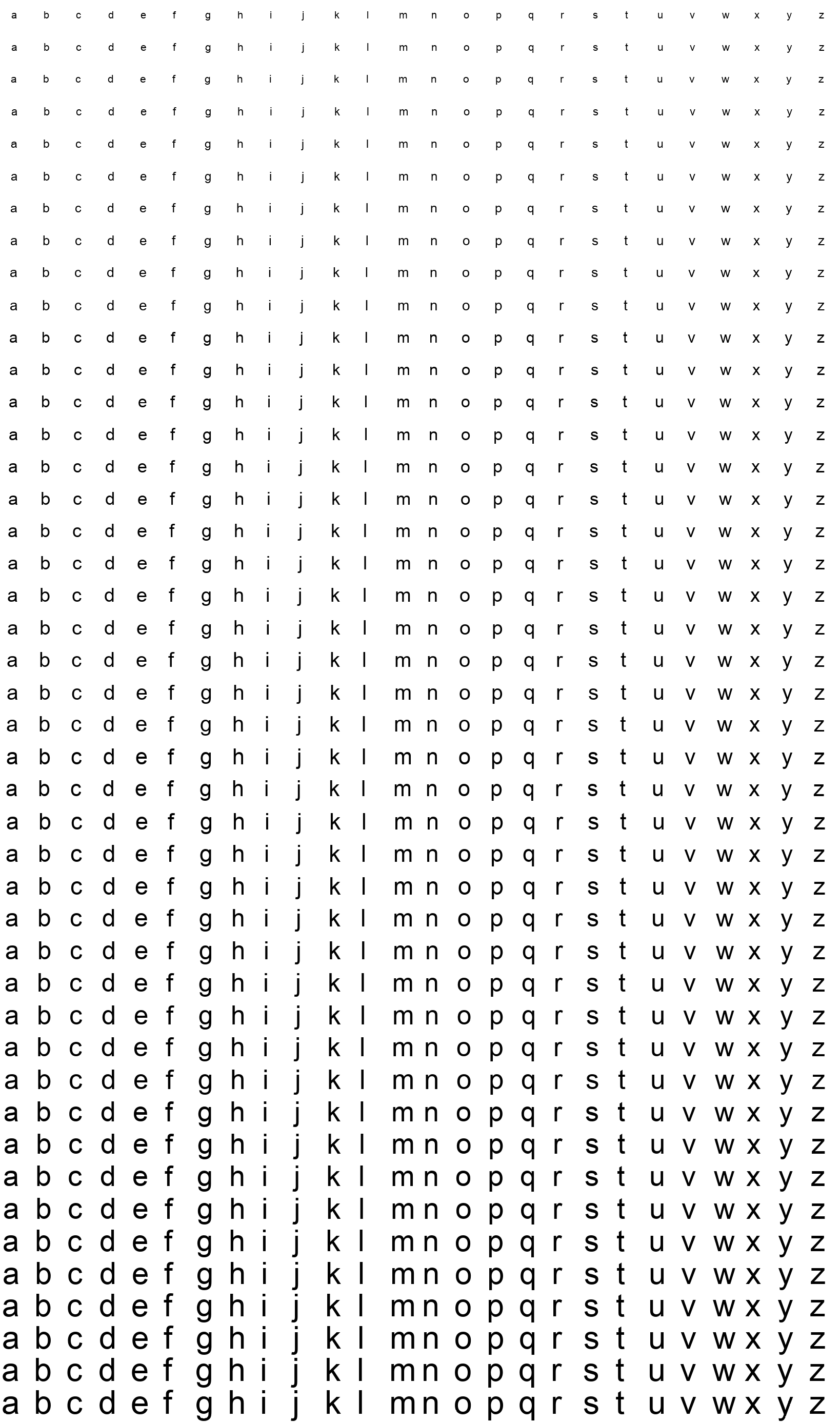}
  \caption{ Arial implicit representation experiment results. All lowercase glyphs are displayed at multiple bitmap sizes (20px - 63px).}
  \label{fig:full-cascade-arial}
\end{figure*}

\begin{figure*}[hbtp]
  \centering
  \includegraphics[width=.7\textwidth]{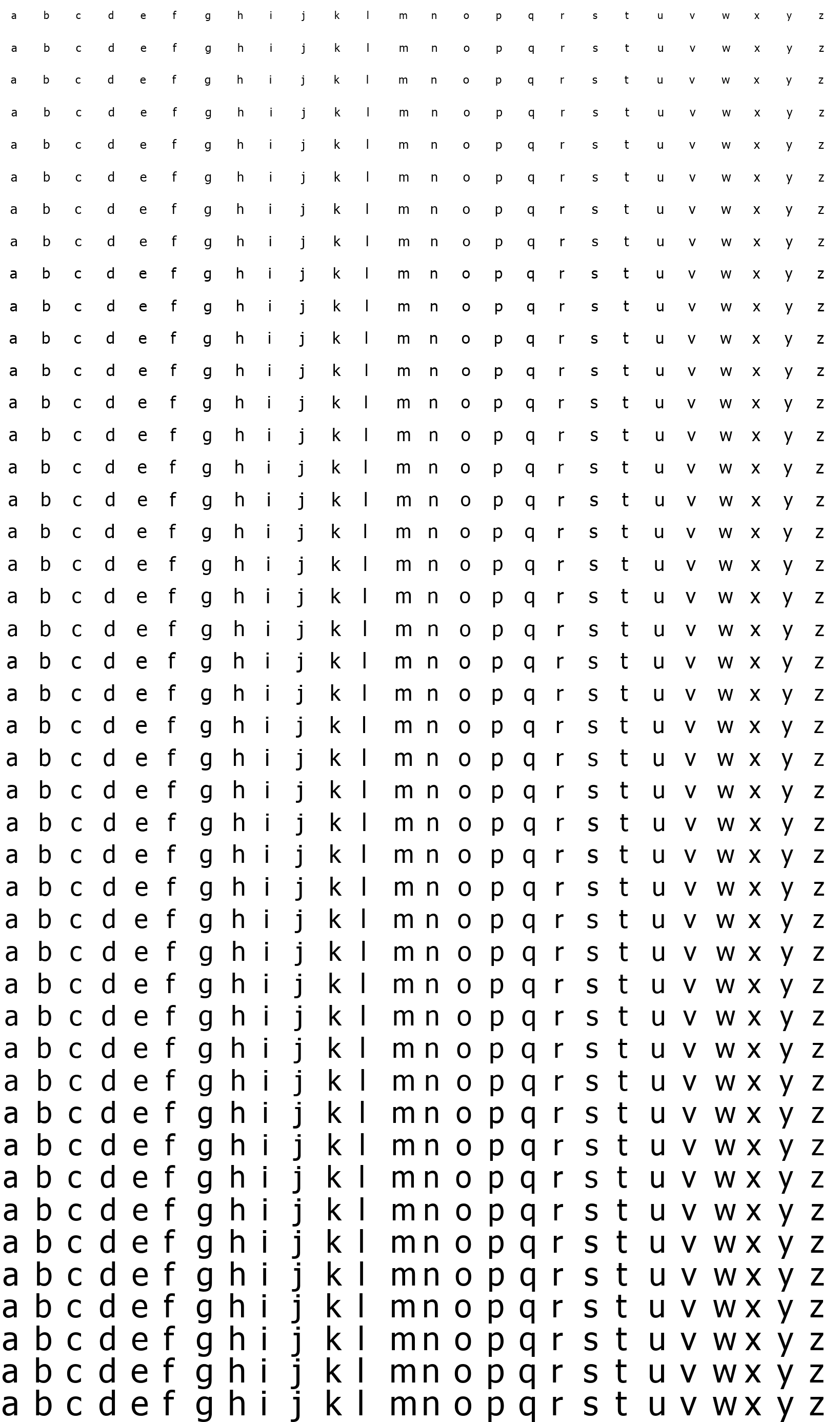}
  \caption{ Tahoma implicit representation experiment results. All lowercase glyphs are displayed at multiple bitmap sizes (20px - 63px).}
  \label{fig:full-cascade-tahoma}
\end{figure*}

\Cref{fig:full-cascade-times-new-roman,fig:full-cascade-arial,fig:full-cascade-tahoma} display all the results for the implicit representation approach on all lowercase glyphs for Times New Roman, Arial, and Tahoma, respectively. The high-quality multi-scale rendering is the essence of this work.

\begin{figure*}[hbtp]
  \centering
  \includegraphics[width=.4\textwidth]{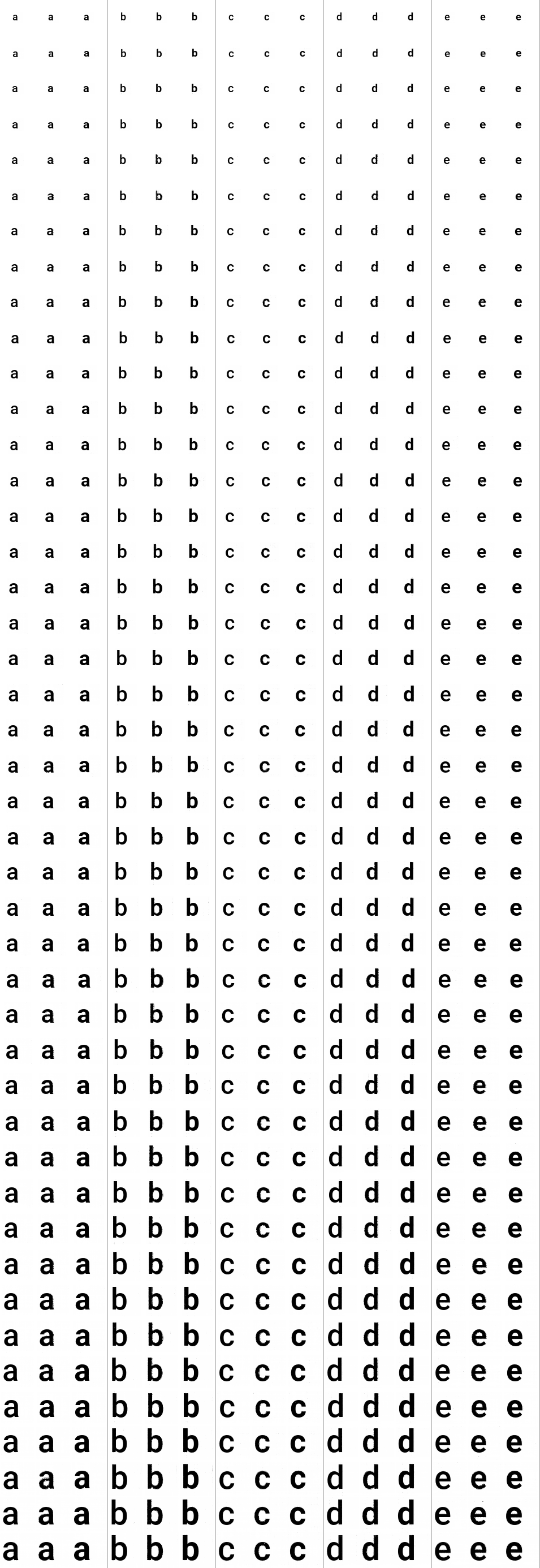}
  \caption{ Roboto weight interpolation for lowercase 'a' to 'e', bitmap sizes 20px - 63px. For each glyph, the ground truth is presented (Medium [left] and Bold [right]), as well as the interpolated weight (center).}
  \label{fig:weight-interpolation-full}
\end{figure*}

\Cref{fig:weight-interpolation-full} displays multi-scale weight interpolation results for lowercase glyphs `a' to `e' for all bitmap sizes between 20px and 63px.

\newpage



\end{document}